\documentclass{article}

\usepackage[eandd, final]{neurips_2026}

\usepackage[utf8]{inputenc} 
\usepackage[T1]{fontenc}    
\usepackage{hyperref}       
\usepackage{url}            
\usepackage{booktabs}       
\usepackage{amsfonts}       
\usepackage{nicefrac}       
\usepackage{microtype}      
\usepackage{xcolor}         
\usepackage{amsmath}
\usepackage{dsfont}
\usepackage{enumitem}
\usepackage{booktabs,tabularx,array}
\usepackage[table]{xcolor}
\usepackage{array}
\usepackage{makecell}

\usepackage{comment}
\usepackage{graphicx}
\usepackage{color}
\usepackage{subcaption} 
\usepackage{geometry}

\usepackage{booktabs}
\usepackage{multirow}

\usepackage{subcaption}
\usepackage{wrapfig}

\newcommand{\name}{\textit{MyoChallenge}}

\title{MyoChallenge 2025: Benchmark Design, Competition Results, and Lessons for Human Athletic Intelligence}
\author{%
  Cheryl Wang$^{1}$\thanks{co-first}
  \quad
  Chun Kwang Tan$^{2*}$
  \quad
  Balint K. Hodossy $^{3*}$
  \quad
  \textbf{Eric Lyu}$^{4}$
  \quad
  \textbf{Jun Guo}$^{5\dagger}$ 
  \and
  \textbf{Wentao Zhao}$^{5\dagger}$ 
  \quad
  \textbf{Huaping Liu}$^{5\dagger}$ 
  \quad
  \textbf{Chengkun Li}$^{6\ddagger}$ 
  \quad
  \textbf{Merkourios Simos}$^{6\ddagger}$ 
  \quad
  \textbf{Bianca Ziliotto}$^{6\ddagger}$ 
  \and
  \textbf{Alexander Mathis}$^{6\ddagger}$ 
  \quad
  \textbf{Siyuan Liu}$^{7\S}$
  \quad
  \textbf{Jiahao Chen}$^{7\S}$
  \quad
  \textbf{Shanlin Zhong}$^{7\S}$
  \quad
  \textbf{Bo Jiang }$^{7\S}$
  \and
  \textbf{Ci Song}$^{7\S}$
  \quad
  \textbf{Yaoye Zhu}$^{5\P}$
  \quad
  \textbf{Chenhui Zuo}$^{5\P}$
  \quad
  \textbf{Yanan Sui}$^{5\P}$
  \quad
  \textbf{Mohamed Irfan Refai} $^{8}$ 
  \and
  \textbf{Massimo Sartori}$^{8}$
  \quad
  \textbf{Guillaume Durandau}$^{1}$
  \quad
  \textbf{Vikash Kumar}$^{9}$
  \quad
  \textbf{Vittorio Caggiano}$^{4}$
  \quad
  \\
  \\
  $^1$McGill University, Canada \quad $^2$Nanyang Technological University, Singapore \and
    $^3$ Imperial College London, UK
  $^4$ King’s College London, UK \and \quad $^5$ Tsinghua University, China 
  $^6$ EPFL, Switzerland $^7$ CASIA, China \and
  $^{8}$University of Twente, Netherlands 
  $^{9}$MyoLab, USA
  \\
  \\[0.6em]
  \small
  $^{\dagger}$Team ActingAI \quad
  $^{\ddagger}$Team Servette MyoClub \quad
  $^{\S}$Team BioSyn \quad
  $^{\P}$Team LNSGroup
  }

\begin{document}

\maketitle

\begin{abstract}
Athletic performance represents the pinnacle of human motor intelligence, demanding rapid choices, precise control, agility, and coordinated physical execution. Although recent advances in robotics have demonstrated remarkable capabilities in replicating complex human movements, these achievements provide limited insight into the underlying neuromuscular coordination that enables such behaviors. Understanding how the nervous system recruits and coordinates muscles during athletic movements remains challenging, as complex muscle activation patterns are difficult to measure directly in vivo, particularly during dynamic, high-intensity activities. To bridge this fundamental gap in understanding, \textit{MyoChallenge} at NeurIPS 2025 established a pioneering benchmark for motor control intelligence in sports, leveraging high-fidelity musculoskeletal models within physics simulation combined with machine learning-driven algorithms. The competition introduced two distinct tracks emphasizing either upper or lower limbs control: a table tennis rally task utilizing a biomechanic upper limb model composed of an arm with a hand (MyoArm) and a trunk (MyoTorso); and a soccer penalty kick using a biomechanic model of legs (MyoLeg) and a trunk (MyoTorso). Marking the fourth iteration of the MyoChallenge series, this event attracted almost 70 teams and over 560 submissions globally, uniting a diverse community ranging from physicians and neuroscientists to machine learning experts. The competition facilitated the development of several state-of-the-art control algorithms for the musculoskeletal system capable of sports agility, leveraging techniques such as physics-based motion planners, on-policy behaviour cloning, hierarchical planning, and muscle synergies. By integrating standardized tasks and physiologically realistic models into the open-source framework of MyoSuite, MyoChallenge'25 serves as a reproducible and reusable testbed to accelerate interdisciplinary research across machine learning, biomechanics, sports science, and neuroscience. Project page: \url{https://www.myosuite.org/myochallenge/myochallenge-2025}

\end{abstract}

\section{Introduction}
\label{sec:intro}

\begin{figure}[h]
\centering
\begin{minipage}[t]{0.49\textwidth}
    \centering
    \includegraphics[width=\textwidth]{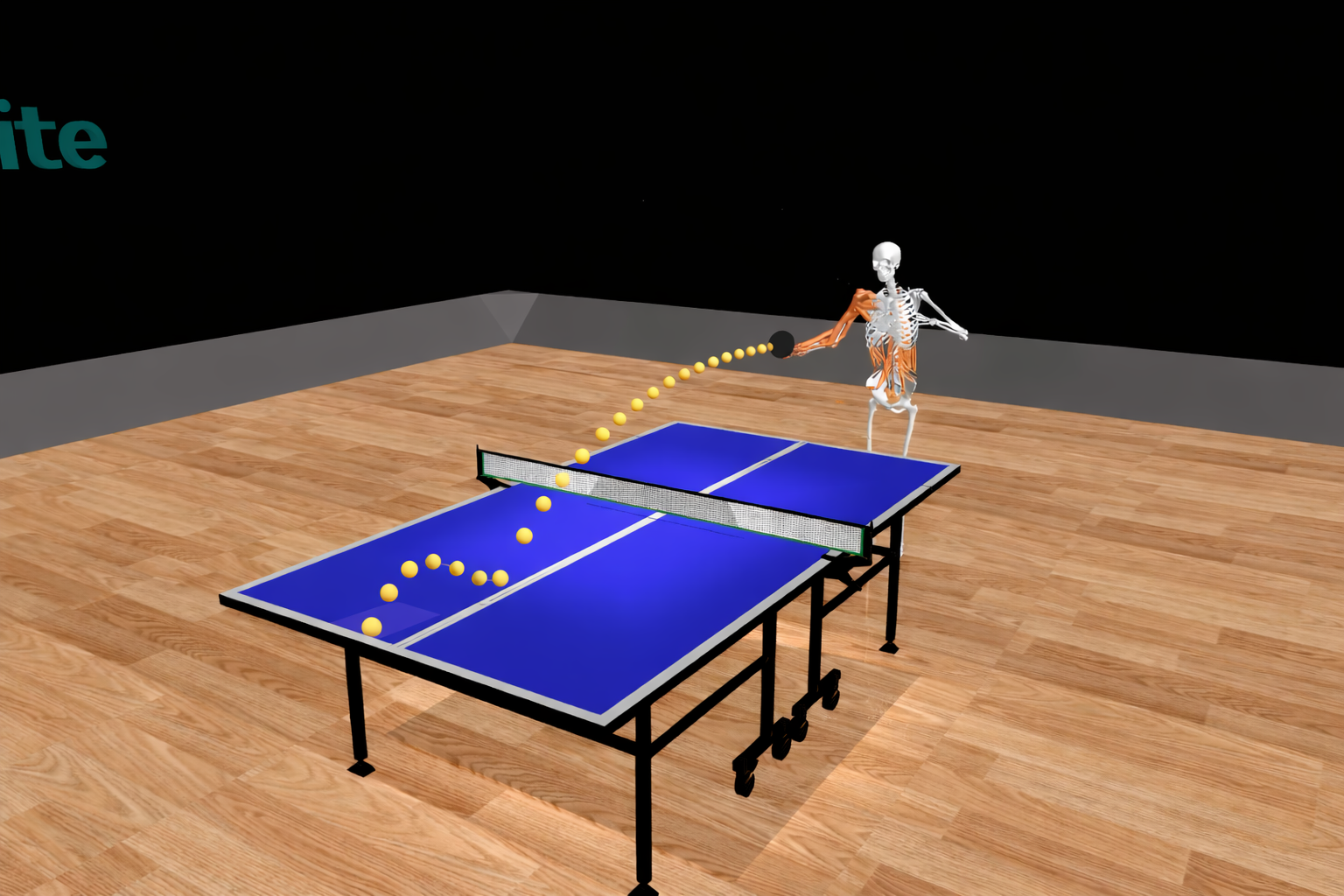}
    \caption*{A}
\end{minipage}\hfill
\begin{minipage}[t]{0.49\textwidth}
    \centering
    \includegraphics[width=\textwidth]{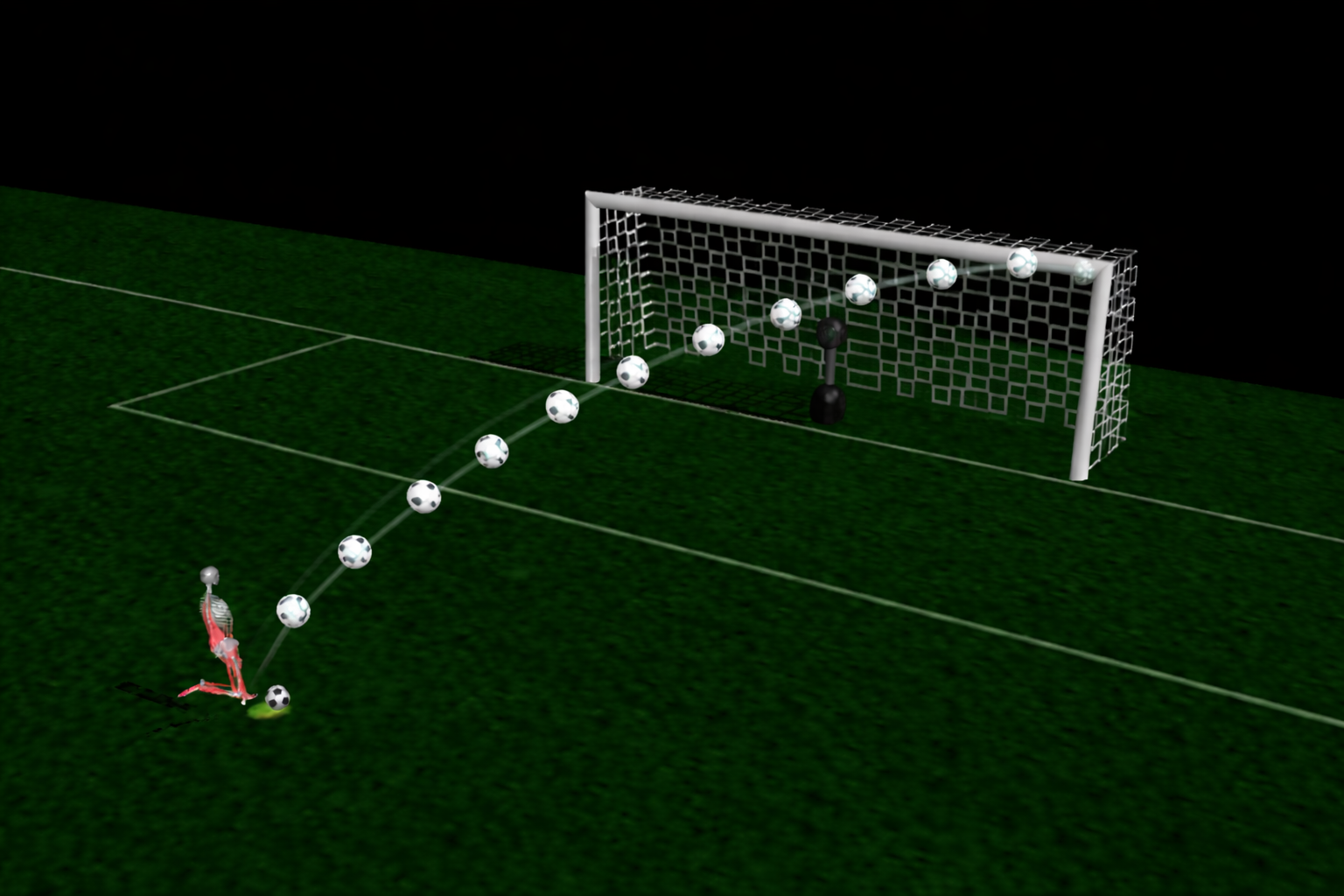}
    \caption*{B}
\end{minipage}

\caption{Two tracks of MyoChallenge 2025. \textbf{A.} The Table Tennis track in which an agent is required to hold a paddle and return an incoming table tennis ball \textbf{B.} The Soccer track where the agent needs to score a penalty goal with or without a goalie.}
\vspace{-0.4cm}
\label{Fig:challenge_tasks}
\end{figure}

Human athletic performance represents one of the most complex manifestations of sensorimotor intelligence, involving continuous control over high-dimensional musculoskeletal (MSk) systems in dynamically changing environments. Understanding human athletic intelligence is a critical pursuit for advancing artificial intelligence systems~\cite{Liu_scirobotics_2022}. Recent advances in humanoid robotics have demonstrated remarkable capabilities in human-like coordination, agile locomotion, and highly dynamic athletic behaviors, including acrobatics and martial arts~\cite{liu2024realdex,bostondynamics2023,He2025ASAP, omniH2O}, navigating uncertain terrain~\cite{zhuang2023}, and collaborating with humans in real time~\cite{zhang2025generative}. 
However, reproducing human-like motion and understanding how such behaviors emerge under physiological constraints remain distinct research objectives. Most humanoid robotic systems rely on simplified actuation models that do not capture the nonlinear muscle--tendon dynamics, delayed force generation, and redundancy characteristic of human MSk systems. Studying athletic behaviors under these constraints is essential for understanding human muscle coordination and adaptive motor control.


Understanding muscle recruitment patterns during athletic movement is also crucial for investigating human motor coordination, recovery, and performance training. Studying athletic behavior under fast, dynamic, and uncertain conditions offers a unique opportunity to understand how the body integrates perception, prediction, and control with applications in prosthetics, rehabilitation, and assistive technologies~\cite{Valero_Cuevas_2024, sartori_neural_2016, Ramdya_Ijspeert_2023, Sartori_2023, song2021deep}. In addition, these studies can enhance biomechanically grounded animation, virtual athletics, and embodied AI agents in simulation environments~\cite{Truong_2024}. Previous MSk research in sports science has largely focused on isolated kinetic performance~\cite{Martin2013} or estimating muscle forces during specific athletic tasks~\cite{Buffi2014,Clancy2023,Attias2026}. Furthermore, these forward simulations are generally restricted to partial body segments and rely heavily on tracking captured MoCap data, limiting their ability to simulate generalized, whole-body sports movements.

In recent years, the community has seen substantial progress across computational biomechanics and ML\cite{ikkala2022breathing,park2022generative,schumacher2023:deprl,chiappa2024acquiring,Kaibo_2024,chiappa2025arnold0}, assistive technologies \citep{Azocar2020,Yu2021,Walia2025}, and physics-based simulation \cite{opensim, IsaacGym_2021, Geijtenbeek2021Hyfydy, caggiano_myosuite_2022}. Within computational biomechanics, several benchmarks have emerged over the past decade to evaluate assistive devices \citep{Kidzinski2019,wang2025myochallenge} and general human motor control \cite{kidzinski2018learning,Song2020,MyoChallenge2022,MyoChallenge2023,caggiano2024myochallenge}. However, no public benchmark currently integrates high-fidelity, full-body MSk models with interactive, physically simulated sports environments. Developing such a benchmark is essential to systematically evaluate the capabilities and limitations of current learning-based approaches under physiologically grounded muscle actuation.

To address this gap, we introduce \textbf{MyoChallenge 2025: Towards Human Athletic Intelligence}, a competition designed to benchmark sports agility using MSk simulations. This work aims to advance the development of realistic biomechanical digital twins of humans, seeking to answer a central question: \textit{Can physiological digital twins achieve human-level coordination during complex sports activities?} The competition features two independent tracks focused explicitly on upper and lower limb coordination. Crucially, these challenges go far beyond static or repetitive motions and demand generalization through reactive, adaptive embodied behaviors, requiring real-time perception-action loops for agile motor control. By providing a reproducible benchmark for complex sports tasks, MyoChallenge aims to advance the development and evaluation of neuromuscular control strategies, with broader implications for biomechanical simulation, rehabilitation, and embodied intelligence.

\definecolor{challengeblue}{RGB}{36,104,160}

\renewcommand{\tabularxcolumn}[1]{%
  >{\raggedright\arraybackslash}m{#1}%
}

\newcolumntype{M}{%
  >{\raggedright\arraybackslash}m{0.17\columnwidth}%
}

\newcolumntype{E}{%
  >{\raggedright\arraybackslash}m{0.18\columnwidth}%
}

\begin{table}[t]
  \caption{Comparison of musculoskeletal models across MyoChallenge years.
  The highlighted rows denote the MyoChallenge~2025 environments introduced
  in this work.}
  \label{tab:myochallenge}

  \centering
  \footnotesize
  \renewcommand{\arraystretch}{1.02}

  \begin{tabularx}{\columnwidth}{
    @{}
    M
    @{\hspace{4pt}}
    c
    @{\hspace{5pt}}
    E
    @{\hspace{4pt}}
    c
    @{\hspace{4pt}}
    c
    @{\hspace{4pt}}
    c
    @{\hspace{5pt}}
    X
    @{}
  }
    \toprule

    \textbf{Model} &
    \textbf{Challenge Year} &
    \textbf{Embodiment} &
    \textbf{Joints\textsuperscript{a}} &
    \textbf{DoFs} &
    \textbf{Muscles} &
    \textbf{Core task} \\

    \midrule

    MyoHand
      & 2022
      & Biological hand
      & 23
      & 23
      & 39
      & Die reorientation and Baoding-ball manipulation \\

    MyoArm
      & 2023
      & Unilateral upper limb
      & 38
      & 27
      & 63
      & Reaching, grasping, and object relocation \\

    MyoLeg
      & 2023
      & Biological lower body
      & 15
      & 28
      & 80
      & Goal-directed chase-tag locomotion \\

    MyoMPL
      & 2024
      & Biological arm with MPL prosthesis
      & 55
      & 44
      & 63\,+\,17\textsuperscript{b}
      & Biological--prosthetic coordination and object handover \\

    MyoOSL
      & 2024
      & Transfemoral limb with OSL prosthesis
      & 17
      & 22
      & 54
      & Prosthetic locomotion over varied terrain \\

    \specialrule{\lightrulewidth}{0pt}{0pt}

    \rowcolor{challengeblue!7}
    \textbf{MyoArm + MyoTorso}
      & \textbf{2025}
      & Upper limb with articulated torso
      & 56
      & 45
      & 273\,+\,2\textsuperscript{c}
      & Moving-ball interception, timed contact, and repeated rallying \\

    \rowcolor{challengeblue!7}
    \textbf{MyoLeg + MyoTorso}
      & \textbf{2025}
      & Lower body with articulated torso
      & 33
      & 46
      & 290
      & Locomotion, stabilization, ball approach, and timed kicking \\

    \bottomrule
  \end{tabularx}

  \vspace{0.5ex}

  \begin{minipage}{\columnwidth}
    \scriptsize
    \textsuperscript{a}Joint counts follow MuJoCo's definition and therefore
    need not equal the number of anatomical joints.
    \textsuperscript{b}Includes 17 prosthetic actuators.
    \textsuperscript{c}Includes two pelvis-position actuators.
  \end{minipage}
  \vspace{-0.6cm}
\end{table}

\section{The MyoChallenge'25 Competition} \label{section2}

For three consecutive years, the MyoSuite team -- a multi-university open-source and open-science initiative to advance human musculoskeletal motor control -- has hosted the MyoChallenge as an official NeurIPS competition to advance the benchmark for controlling human digital twins using frontier machine learning algorithms. Marking its fourth edition, MyoChallenge~'25 introduces two athletic-intelligence tracks that substantially increase musculoskeletal and task complexity while retaining separate focus on upper-limb control and locomotion (Table~\ref{tab:myochallenge}). Whereas previous editions used models with at most 80 actuators, the 2025 Table Tennis and Soccer tasks require control of 275 and 290 actuators, respectively. The Table Tennis track extends previous upper-limb manipulation tasks from object reorientation and relocation to grasping and orienting a paddle, intercepting a moving ball, and directing its return, which requires precise contact timing and fluent upper-body control. The Soccer track similarly extends previous lower-limb locomotion tasks by requiring coordinated approach, balance, and a precisely timed kick against a moving goalkeeper. To keep these added demands focused on the intended control challenges, pelvis translation was position-actuated in Table Tennis, while the arms were omitted in Soccer.


The competition was divided into an opening phase and a playoff phase. In the initial phase, submissions were evaluated using static environments with no task variations, encouraging participants of all skill levels to test early solutions. Teams scoring above 10\% advanced to the playoff phase, which introduced significant environmental variations to test model robustness. 
The following sections outline the simulation environment and musculoskeletal models (Sec. \ref{sec:MSK}), the proposed tasks and evaluation criteria (Sec. \ref{sec:Task&Env}), and the results of the competition (Sec. \ref{sec:results}).

\subsection{Simulation Environment, Models, and Data} \label{sec:MSK}

\paragraph{Simulation Framework} The competition and the complete set of musculoskeletal models are embedded within MyoSuite, an open-source collection of environments and tasks powered by the MuJoCo physics engine \citep{todorov2012mujoco}. MyoSuite provides physiologically realistic musculoskeletal models \cite{myosim} within a simulation framework that significantly surpasses the speed of state-of-the-art simulators \cite{ikkala2020,Erez_2015} used in prior competitions, such as OpenSim. Furthermore, MyoSuite natively supports comprehensive contact dynamics, enabling the simulation of complex, contact-rich environment interactions. Within MuJoCo, the muscles are specifically modelled as actuators attached to tendons, which are assumed to be infinitely stiff and without elasticity.

\paragraph{Musculoskeletal Models}
Each model segment used in the two tasks is presented in detail below:
\begin{itemize}[noitemsep, nolistsep, leftmargin=*]
    \item \textbf{MyoArm} The right arm of the MyoArm model, comprising 27 degrees of freedom (DOFs) and 63 muscle-tendon units, is adapted from the OpenSim MoBL arm model \cite{saul2015benchmarking,www_mobl}. This model has been utilized in prior MyoChallenges for manipulation tasks \cite{MyoChallenge2022,MyoChallenge2023,wang2025myochallenge}.
    \item \textbf{MyoTorso} The MyoTorso is created based on \cite{Christophy2011} in \citep{Walia2025} with 210 actuators and 18 DoFs, and has been used in previous balancing research \citep{Wang2025}.
    \item \textbf{MyoLeg} MyoLeg model comprises 80 muscle-tendon and 28 DoFs units derived from the OpenSim lower limb model \cite{Rajagopal2016} and was previously featured in the MyoChallenges \cite{MyoChallenge2023, wang2025myochallenge}.   
\end{itemize}

\paragraph{Data} Participation in the competition did not require pre-collected training datasets. The primary challenge was to develop policies that learn directly through interaction and exploration within the \textit{MyoSuite} environments. However, to support alternative approaches like imitation learning and behavior cloning, we provide a custom MyoChallenge bot (Appendix \ref{appendix:myochallenge_bot}) that allows participants to generate and record their own training data. We also encouraged the use of public motion capture datasets, such as AMASS \cite{AMASS:ICCV:2019}. Following the conclusion of the competition, further relevant datasets have become available, e.g., the BONES-SEED dataset \cite{bones_seed_2026}. While optional, these datasets can serve as strong motion priors for imitation learning or behavior cloning.

\subsection{Tasks and Evaluations}\label{sec:Task&Env}

\subsubsection{Table Tennis Track}

\textbf{Task} The table tennis track (Fig.\ref{Fig:challenge_tasks}-A) features a single-arm and trunk task where the agent must hold on to a table tennis paddle and return an incoming ball trajectory, while controlling the torso posture. To simplify the task, the pelvis is given two position actuators in the horizontal x and y directions to allow positioning around the table. The left arm is left unactuated and is fixed in a static position to help with balance. The paddle is initiated close to the right palm in an ungrasped position. A key source of complexity in this task is that a generalist policy that must swiftly handle the control across incoming table tennis ball trajectories with various speeds. The environmental variations across different phases and evaluations are detailed in Appendix \ref{appendix:env} - Table \ref{table:task_object_properties} and the variations in Table \ref{table:tt_phase_comparison}.

\textbf{Observation} The controller receives a 416-dimensional observation vector, including MSk joint positions and velocities, muscle stimulation levels, and the 6-DoF positions and velocities of the table tennis ball and paddle. Additionally, contact status is indicated by six binary labels specifying which object is the ball in contact with, if any (see Appendix \ref{app:tt} -  Table \ref{table:tennis_observation_space} for details). 

\textbf{Action} The action space includes a 275-dimensional continuous vector ranging from [-1, 1]. This includes the MyoArm, consisting of 63 muscles, the MyoTorso, consisting of 210 muscles and two position actuators for pelvis translation in the x,y plane. The actions targeting muscles are then renormalized to a [0, 1] range within the environment, to be interpreted as muscle excitation levels.

\textbf{Termination}  A trial is successful if the paddle strikes the ball exactly once, sending it directly to the opponent's side without touching the agent's half of the table, or missing the table entirely. The termination condition can be found in full in Appendix \ref{app:tt}.

\textbf{Evaluation Metric} The participants were first ranked based on the number of successful returns of table tennis. During phase 2, teams that scored within 10\% of each other in the 'success rate' metrics were then re-ranked based on the second criterion, which is the muscle activation effort, with less effort leading to a better ranking. 

\textbf{Baseline Controller} We provided a benchmarking baseline policy that used a heuristic muscle grouping dimensional reduction strategy presented in \cite{Wang2025} and further trained with a deep neural network with reinforcement learning. The baseline demonstrated a success rate of approximately 2\% under Phase 1 variation. Detailed hyperparameters can be found in Appendix \ref{app:tt}.

\subsubsection{Soccer Track}

\textbf{Task} The objective is to develop a controller for the fullbody MSk model that allows for coordinated locomotion and kicking of a ball to score goals in a net with and without a goalkeeper, as shown in Fig.\ref{Fig:challenge_tasks}-B. To reduce the number of muscles, the model has no arms. A key source of complexity in this task is coordinating 290 muscles to produce a single movement sequence that requires the legs to generate locomotion, maintain postural stability, and deliver a precisely targeted ballistic strike. These functions place competing demands on the same effectors. The environmental variations and goalkeeper behaviour are detailed in Appendix \ref{app: soccer} - Table \ref{table:task_object_properties_soccer} and the variations in Table \ref{table:soccer_phase_comparison}. 

\textbf{Observation} The controller/policy has access to a 418-dimension proprioceptive vector that includes information such as joint angles, velocities, muscle states, and goal conditions, including soccer, keeper and net, updated at each time step (Appendix \ref{app: soccer} - Table \ref{tab:obs_space_soccer}).

\textbf{Action} The action space consists of 290 continuous muscle excitation controls ranging from  [-1, 1].

\textbf{Termination} A simulation trial is deemed complete when the player (or agent) scores the ball such that it is fully within the confines of the net within the maximum time constraints stated in Table \ref{table:soccer_phase_comparison}.

\textbf{Evaluation Metrics}. The performance was first measured by the success rate of scoring goals. During phase 2, teams that scored within 10\% of each other in the 'success rate' metrics were then ranked based on the muscle activation effort, with less effort leading to a better ranking. 

\textbf{Baseline Controller} We did not provide a baseline controller for the soccer task. However, baseline controllers for the intact myoLeg model that produce walking are available in MyoSuite \citep{caggiano_myosuite_2022}, such as DEP-RL controller \citep{schumacher2023:deprl} and a reflex-based controller \citep{song2015neural}. 

\section{Benchmark and Competition Results}
\label{sec:results}

\subsection{Performance of Standard RL Baselines}

To assess whether existing off-the-shelf reinforcement learning methods are sufficient for the proposed tasks, we evaluate PPO and SAC together with muscle-grouping and curriculum-learning variants. All methods are trained for 10 million environment steps using the default Stable-Baselines3 hyperparameters and evaluated over 100 unseen Phase~1 trials.  In addition to task success and effort, we report intermediate behavioral competencies to identify where each method fails.

\definecolor{baselineblue}{RGB}{232,241,248}

\newcommand{\cmark}{%
  \textcolor{green!45!black}{\ensuremath{\checkmark}}%
}
\newcommand{\xmark}{%
  \textcolor{red!70!black}{\ensuremath{\boldsymbol{\times}}}%
}

\newcolumntype{Y}{%
  >{\raggedright\arraybackslash}X%
}
\newcolumntype{C}[1]{%
  >{\centering\arraybackslash}p{#1}%
}

\begin{table}[t]
  \caption{Phase~1 evaluation results over 100 trials for the Table Tennis and Soccer tracks. All methods were trained for 10M steps using Stable-Baselines3 with default hyperparameters. The shaded row denotes the provided Table Tennis baseline; no baseline was provided for the Soccer track.}
  \label{tab:phase1-results}

  \centering
  \small
  \setlength{\tabcolsep}{4pt}
  \renewcommand{\arraystretch}{1.05}

  \begin{tabularx}{\columnwidth}{
    @{}
    Y
    C{0.14\columnwidth}
    C{0.14\columnwidth}
    C{0.20\columnwidth}
    C{0.11\columnwidth}
    C{0.06\columnwidth}
    @{}
  }
    \toprule

    \multicolumn{6}{@{}l}{
      \textbf{(a) Table Tennis Track}
    } \\

    \addlinespace[0.35ex]

    \textbf{Algorithm/Method} &
    \mbox{\bfseries Holding paddle} &
    \mbox{\bfseries Maintain upright} &
    \mbox{\bfseries R-Arm Upright} &
    \mbox{\bfseries Success rate} &
    \mbox{\bfseries Effort} \\

    \midrule

    \rowcolor{baselineblue}
    \textbf{PPO + muscle grouping + CL (Baseline)}
      & \cmark
      & \cmark
      & \cmark
      & 2\%
      & 0.79 \\

    PPO + muscle grouping
      & \cmark
      & \cmark
      & \xmark
      & 0\%
      & 0.20 \\

    PPO
      & \cmark
      & \cmark
      & \xmark
      & 0\%
      & 0.51 \\

    SAC + muscle grouping
      & \cmark
      & \cmark
      & \xmark
      & 0\%
      & 0.67 \\

    SAC
      & \xmark
      & \xmark
      & \xmark
      & 0\%
      & 0.63 \\
 \end{tabularx}

  \begin{tabularx}{\columnwidth}{
    @{}
    Y
    C{0.14\columnwidth}
    C{0.14\columnwidth}
    C{0.20\columnwidth}
    C{0.11\columnwidth}
    C{0.06\columnwidth}
    @{}
  }
    \toprule

    \multicolumn{6}{@{}l}{
      \textbf{(b) Soccer Track} \textit{(no baseline)}
    } \\

    \addlinespace[0.35ex]

    \textbf{Algorithm/Method} &
    \mbox{\bfseries Torso upright} &
    \mbox{\bfseries Walking forward} &
    \mbox{\bfseries Kicking Soccer} &
    \mbox{\bfseries Success rate} &
    \mbox{\bfseries Effort} \\

    \midrule

    PPO + muscle grouping
      & \xmark
      & \xmark
      & \xmark
      & 0\%
      & 0.09 \\

    PPO
      & \xmark
      & \xmark
      & \xmark
      & 0\%
      & 0.17 \\

    SAC + muscle grouping
      & \xmark
      & \xmark
      & \xmark
      & 0\%
      & 0.33 \\

    SAC
      & \xmark
      & \xmark
      & \xmark
      & 0\%
      & 0.32 \\

    \bottomrule
  \end{tabularx}
  \vspace{-0.75cm}
\end{table}

\begin{table}[t]
  \centering
  \caption{Ranking results for the two tracks of MyoChallenge 2025.
  Effort is recomputed as mean squared activation with a reward
  weight of $1$; see Appendix~\ref{appendix:env}.}
  \label{table:results}

  \normalsize
  \setlength{\tabcolsep}{3pt}
  \renewcommand{\arraystretch}{1.0}

  \begin{tabularx}{\columnwidth}{
    @{}
    l
    >{\raggedright\arraybackslash}X
    r
    r
    >{\raggedright\arraybackslash}X
    r
    r
    @{}
  }
    \toprule
    & \multicolumn{3}{c}{\textbf{Table Tennis Track}}
    & \multicolumn{3}{c}{\textbf{Soccer Track}} \\
    \cmidrule(lr){2-4}
    \cmidrule(l){5-7}

    \textbf{Rank}
      & \textbf{Team}
      & \textbf{Score (\%)}
      & \textbf{Effort}
      & \textbf{Team}
      & \textbf{Score (\%)}
      & \textbf{Effort} \\
    \midrule

    \multicolumn{7}{c}{\textit{Phase 1}} \\
    \midrule

    1st Place
      & ActingAI
      & 100
      & 0.3277
      & Servette MyoClub
      & 100
      & 0.9265 \\

    2nd Place
      & CIAO Group
      & 96
      & 0.7428
      & Ligament Legends
      & 0
      & 0.2393 \\

    3rd Place
      & Servette MyoClub
      & 94
      & 0.8229
      & N/A
      & N/A
      & N/A \\

    \midrule
    \multicolumn{7}{c}{\textit{Phase 2}} \\
    \midrule

    1st Place
      & ActingAI
      & 64
      & 0.1306
      & Servette MyoClub
      & 32
      & 1.3174 \\

    2nd Place
      & BioSyn
      & 60
      & 0.1297
      & N/A
      & N/A
      & N/A \\

    3rd Place
      & LNSGroup
      & 61
      & 0.8473
      & N/A
      & N/A
      & N/A \\

    \bottomrule
  \end{tabularx}
  \vspace{-0.4cm}
\end{table}

Table~\ref{tab:phase1-results} shows that standard, off-the-shelf, model-free RL fails to solve either task within the evaluated training budget. In Table Tennis, the PPO variants and SAC with muscle grouping learn basic paddle retention and postural stability, but only the curriculum-learning baseline achieves correct arm positioning and any successful returns, with a success rate of just 2\%. In Soccer, none of the tested methods acquires even the prerequisite behaviors of standing upright. These results expose the exploration and coordination limitations of off-the-shelf methods in high-dimensional musculoskeletal systems and motivate the development of new structured solutions through the competition.

\subsection{Competition Results}
This year's \name{} saw a total of 69 participating teams from over $10$ countries. Over the four-month submission period, we received a total of $562$ submissions across two phases and recorded over 16k total downloads of MyoSuite. Among the teams that completed the post-competition survey, $63\%$ were composed entirely of students, including four teams of bachelor's students. The DEI (Diversity, Equity, and Inclusion) award was presented to Team KAIST\_RISE from KAIST, Korea, which featured women researchers in physiology and became the first team without a machine learning background to achieve a score of 41\% in Phase 1. The Student Award was given to MusclePower, a team consisting of a single master's student from Imperial College London, UK, which achieved a score of up to 18\% in Phase 2. However, no participants came from South America, Oceania, or Africa, suggesting that further extension is needed within these underrepresented communities.

This year's table tennis track attracted the highest number of competing teams since MyoChallenge 2022. In phase 1, more than 12 teams scored beyond baseline, with 7 of them scoring more than $90 \%$; in phase 2, 9 teams competed for the final winner.  The winning team in the table tennis achieved 100\% in phase 1 and a 64\% success rate in phase 2 with an effort of 0.1306 (average muscle activation units). The top three teams underwent a fierce competition of only a $5\%$ difference in success rate in the final scoring. Due to the complexity of the soccer task, only one team was able to surpass the provided baselines during the period of competition (Table \ref{table:results}). To further advance development in sports biomechanics, the competition suite will be retained within the MyoSuite repository.

\subsection{Table Tennis Winning Solutions} 

\subsubsection{First Place - Team ActingAI}

\begin{figure}[h]
    \centering
    \includegraphics[height=6.5cm]{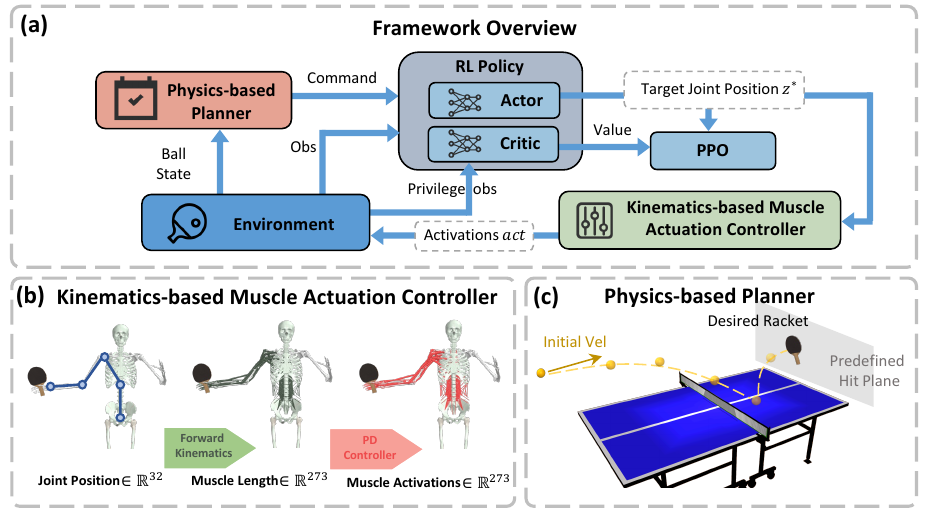}
    \caption{ (a) \textbf{Framework overview.} The policy takes as input current observation and high-level commands from the planner, and generates the target joint positions, which are subsequently translated into muscle signals through the Kinematics-based Muscle Actuation Controller. (b) \textbf{Kinematics-based Muscle Actuation Controller.} Through proving the differential flatness, Diff-Muscle integrates forward kinematics and a PD controller to translate target joint positions into muscle activations. (c) \textbf{Physics-based Planner.} Given the ball state, the planner predicts the desired racket position, orientation, and velocity at the predefined plane as high-level commands.  }
    \label{fig1} 
    \vspace{-0.5cm}
\end{figure}

\paragraph{Architecture Overview} ActingAI propose \textbf{Diff-Muscle}, a hierarchical reinforcement learning framework designed based on differential flatness to enable dimensionality reduction and efficient control for complex musculoskeletal systems in the highly dynamic table tennis task. As illustrated in Figure~\ref{fig1}(a), the framework consists of a high-level Physics-based Planner, a low-level RL Policy, and a Kinematics-based Muscle Actuation Controller (K-MAC). As illustrated in Figure~\ref{fig1}(c), the planner predicts the incoming ball's trajectory and computes target racket states (position, velocity, and orientation) as commands. The RL policy, optimized via Proximal Policy Optimization (PPO), generates target joint positions based on these commands. These joint targets are then translated into physiologic muscle activation signals through the K-MAC module.

\paragraph{Key Solution Insight} ActingAI's key insight is that musculoskeletal systems, despite their complexity, exhibit \textbf{conditional differential flatness}~\cite{sira2018differentially,zhao2026diff} with respect to joint configurations. Through rigorous mathematical proof, the group demonstrates that all system states and control inputs can be algebraically determined by a set of flat outputs (the joint angles $z$) and their finite derivatives. As illustrated in Figure~\ref{fig1}(b), this theoretical foundation allows ActingAI to transform the policy exploration space from the high-dimensional muscle activation space ($\mathbb{R}^{273}$ for our MyoSuite model) to a significantly lower-dimensional joint space ($\mathbb{R}^{32}$), effectively compressing the action space by nearly an order of magnitude without loss of dynamical information. Their proposed K-MAC establishes a principled bridge between joint-level planning and muscle-level execution through kinematics priors, analytically mapping target joint configurations to physiological muscle activations. This ensures that the learned policy explores a structurally valid manifold, bypassing the inefficient search in the null space of redundant muscle forces. By utilizing kinematic priors, Diff-Muscle achieves higher exploration efficiency and lower energy consumption compared to end-to-end RL or learned synergy baselines.

\subsubsection{Second Place - Team BioSyn}
\paragraph{Architecture Overview} 

Team BioSyn used a hybrid framework combining hierarchical planning and structured reinforcement learning. In the table tennis task, trajectory generation and stroke execution were guided by simplified models of ball flight and paddle-ball interaction \cite{su2025hitter}. The structured reinforcement learning module explicitly modelled muscle synergies by assigning specialized subnetworks to different body regions, and leveraged synergetic exploration to improve search efficiency in the redundant muscle space \cite{liu2026biosyngrasp}. Furthermore, a weakness-aware curriculum learning strategy was introduced to emphasize underperforming stroke patterns during the later phase of training, leading to an overall performance gain.

\begin{figure}[h]
  \centering
  \includegraphics[width=0.8\linewidth]{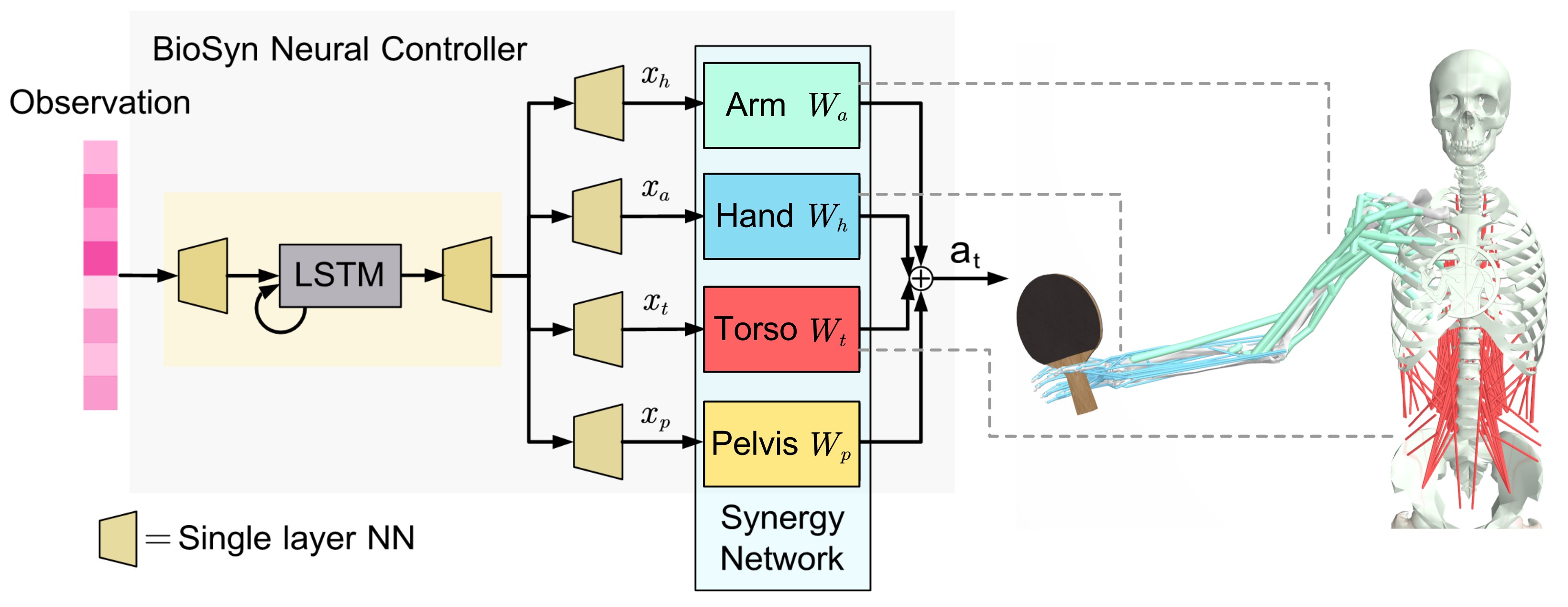}  
  \caption{BioSyn neural controller. The controller is structured according to the musculoskeletal system, with synergy subnetworks dedicated to the hand, arm, torso, and pelvis.}
  \label{fig:biosyn}
\end{figure}

\paragraph{Key Solution Insight - Structured Reinforcement Learning}
BioSyn was initially inspired by evidence that different effectors are associated with distinct control regions in the human motor cortex. Based on this insight, muscle control was decomposed across different body parts rather than learned through a single monolithic controller. As illustrated in Fig.~\ref{fig:biosyn}, the neural network was designed to reflect the structure of the musculoskeletal system, with synergistic subnetworks dedicated to the hand, arm, torso, and pelvis. Furthermore, to improve learning efficiency in the high-dimensional muscle action space, the weights of these synergy networks were incorporated into the exploration process via multivariate Gaussian noise (see Appendix \ref{app:biosyn} for full equation). Further analysis in \cite{liu2026biosyngrasp} showed that BioSyn induces more correlated control patterns among functionally similar muscles, thereby reducing antagonistic co-activation and improving muscular energy efficiency.

\subsubsection{Third Place - Team LNSGroup}

\paragraph{Architecture Overview} To address the challenges of the complex table tennis task, the LNSGroup team proposed a framework integrating physical priors, CrossQ~\cite{bhattcrossq}, and a Gaussian Process Classifier (GPC)~\cite{williams2006gaussian} hybrid strategy. A data-calibrated physical model predicts the incoming ball's trajectory in real time, yielding explicit spatiotemporal targets that reshape dense reward signals. For low-level actuation, the CrossQ algorithm ensures stable high-dimensional control. At the high level, a GPC selects between specialized forehand and backhand policies based on incoming-ball features and estimated policy success probabilities.

\begin{figure}[htbp]
    \centering
    \includegraphics[width=0.95\textwidth]{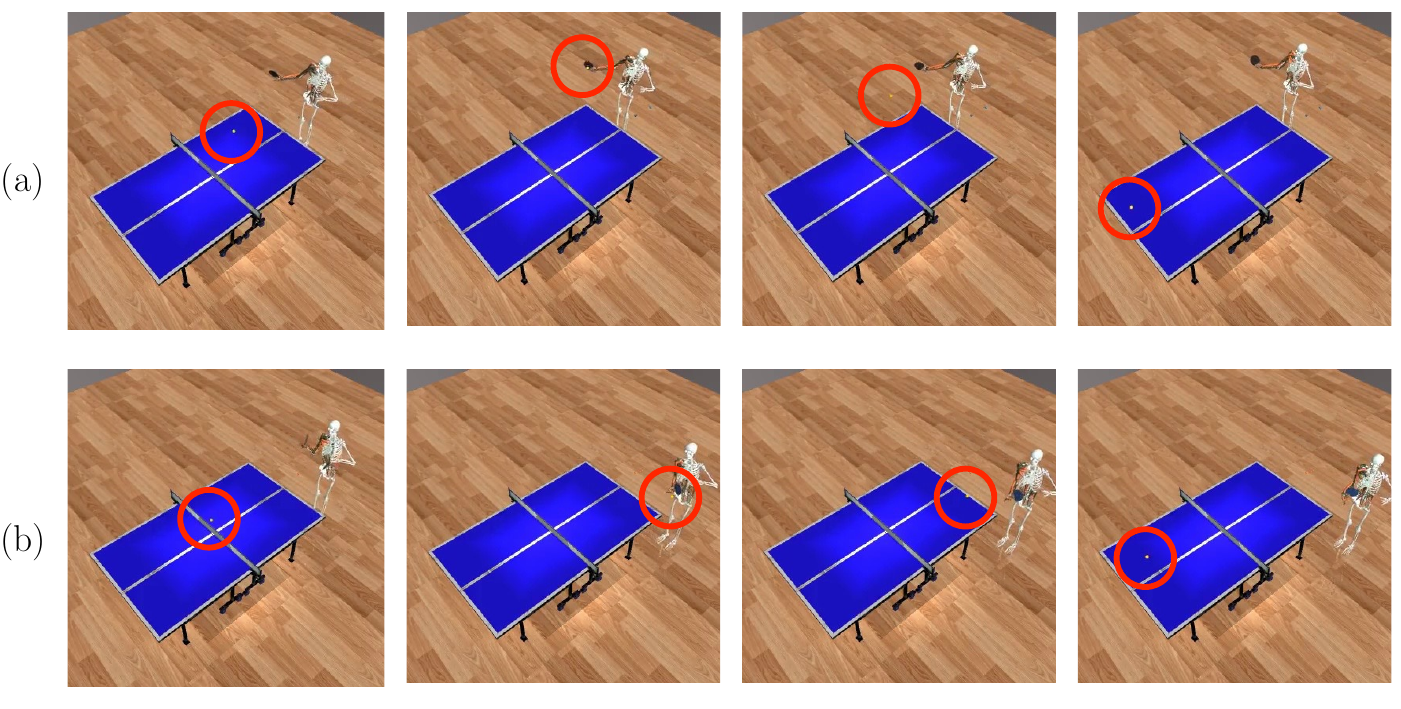}
    \caption{Visualization of the agent executing (a) a forehand strike and (b) a backhand strike.}
    \label{fig:strokes}
    \vspace{-0.4cm}
\end{figure}

\paragraph{Key Solution Insights}

LNSGroup's key insight was to reduce open-ended muscle control into confidence-weighted stroke execution around physics-derived targets. A lightweight ball-flight model, calibrated by trajectory sampling~\cite{su2025hitter}, predicted the hitting position, paddle orientation, velocity, and strike time, turning sparse returns into dense geometric tracking rewards; rule-based pelvis control stabilized the torso so RL could focus on upper-limb striking. The low-level controller used CrossQ~\cite{bhattcrossq}, an off-policy actor-critic method that removes target networks and instead evaluates current and next state-action pairs in a shared batch. This design uses batch normalization effectively under end to end training, improving sample efficiency and training robustness for high-dimensional actions. To cover the broadened Phase-2 trajectory distribution, LNSGroup trained specialized forehand and backhand policies and used a GPC~\cite{williams2006gaussian} to estimate each policy's success probability from incoming-ball features, selecting the stroke with the highest confidence at runtime.

\subsection{Soccer Penalty Winning Solution}

\subsubsection{First Place - Servette MyoClub}

\paragraph{Architecture Overview} Rather than training a controller from scratch, which would require jointly solving locomotion, balance, and kicking, Servette MyoClub took a more principled approach by building on an existing policy, KINESIS ~\cite{simos2025kinesis}, that already produces human-like gait via imitation learning, and that had been extended to support goal-directed movement. KINESIS was trained via motion imitation on 1.8 hours of human locomotion data (KIT-Locomotion dataset~\cite{mandery2015kit}), outputting muscle activations for a full-body model with up to 290 muscles. The architecture comprises a Mixture of Experts (MoE) with 3 expert policies (deep MLPs with 6 hidden layers of sizes \([2048, 1536, 1024, 1024, 512, 512]\), SiLU activations) and a gating network for expert selection. Each expert maps proprioceptive and target observations (pelvis state, joint kinematics, foot contacts, target reference pose) to high-dimensional muscle activation signals. For the challenge solution, Expert~1, the most extensively trained expert (10{,}000 epochs) which is specialized in forward locomotion~\cite{simos2025kinesis}, was used as the foundation for the soccer adaptation pipeline described below.

\begin{figure*}[h!]
    \centering
    \vspace{-0.3cm}
    \includegraphics[width=1\linewidth]{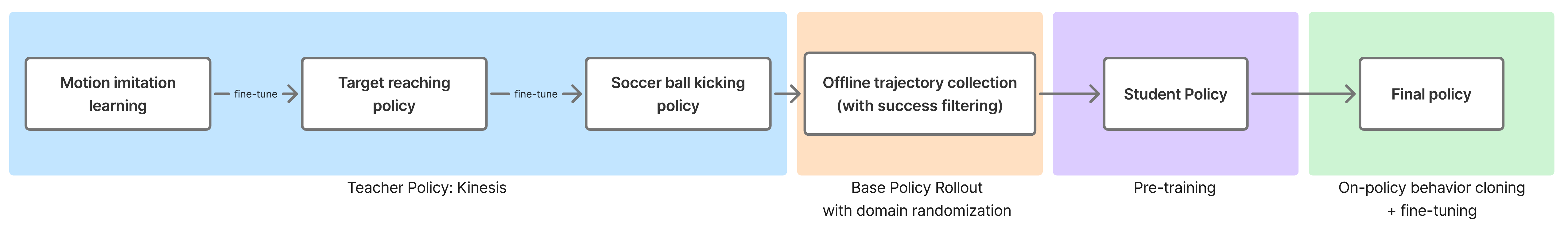}
    \caption{Overview of Servette MyoClub training pipeline.}
    \label{fig:pipeline}
    \vspace{-0.4cm}
\end{figure*}

\paragraph{Key Solution Insight}

\begin{wrapfigure}{r}{0.5\linewidth}
    \centering
    \vspace{-1.5em}
    \includegraphics[width=1\linewidth]{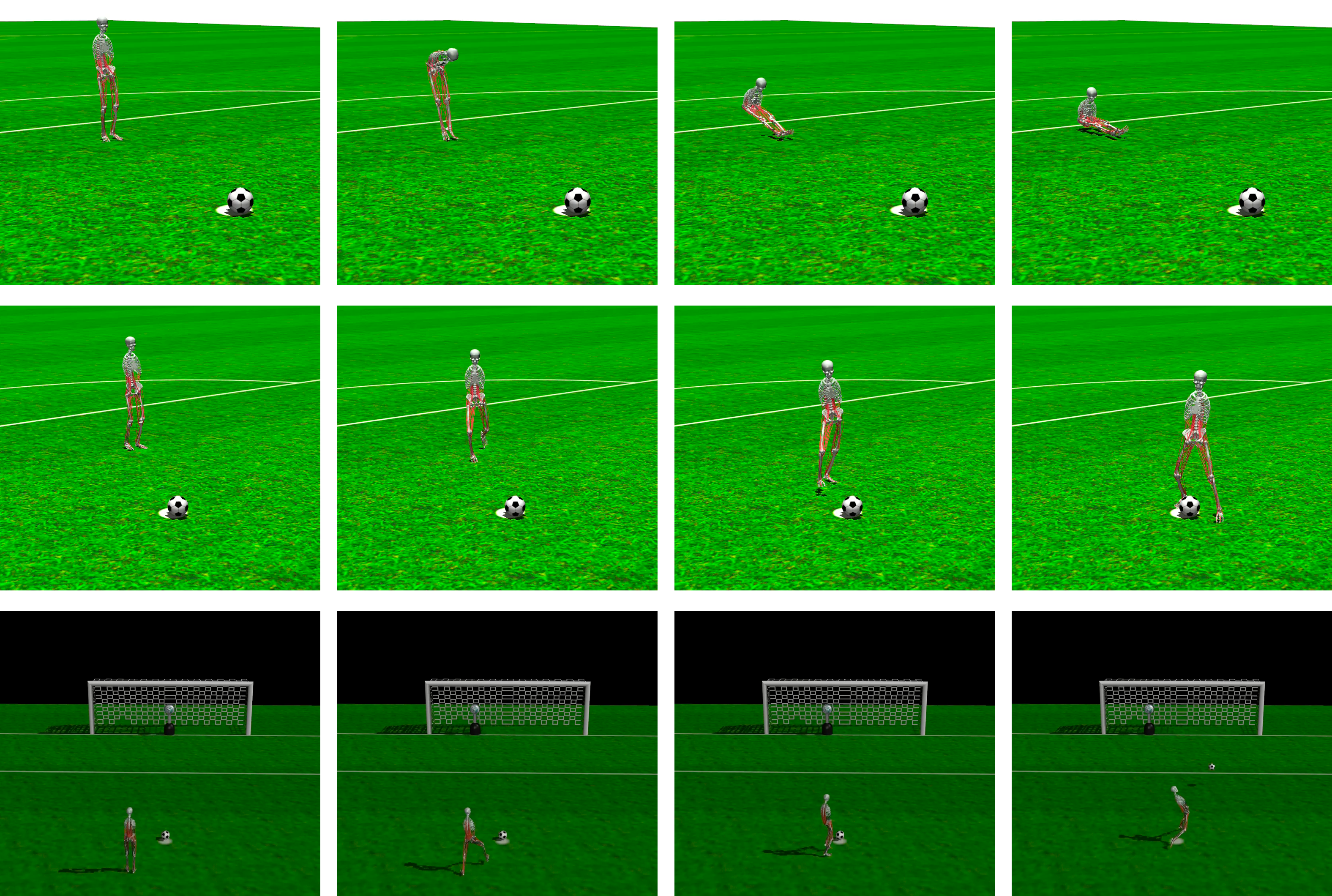}
    \caption{Task visualization. \textit{Top}: Ball-reaching performance at from-scratch initialization. \textit{Middle}: Zero-shot ball-reaching performance by KINESIS. \textbf{Bottom}: Final ball kicking performance after on-policy behavior cloning.}
    \label{fig:viz}
    \vspace{-2em}
\end{wrapfigure}

The winning solution decomposed the penalty kick task into two stages (see Figure~\ref{fig:pipeline} for the full pipeline):

\begin{enumerate}[leftmargin=*, label=\arabic*.]
  \item \textbf{Teacher policy development}. Starting from the KINESIS motion imitation policy (Expert~1), a soccer ball-kicking policy was derived through two successive fine-tuning steps: first into a target-reaching policy, then into a kicking policy. For convenience, this stage operated entirely within the KINESIS observation space.

  \item \textbf{Retargeting to the MyoChallenge environment.} Because the KINESIS and MyoChallenge environments define different observation spaces, the ball-kicking policy from stage~1 was transferred into a student policy that operated on MyoChallenge observations through offline trajectory collection, student pre-training, on-policy behavior cloning~\cite{chiappa2025arnold0}, and reinforcement learning fine-tuning. 
\end{enumerate}


\section{Discussion} \label{sec:dis}

\paragraph{Generalizable Design Principles}
Across both MyoChallenge 2025 tracks, no single RL algorithm, action representation, or reward formulation provided a plug-and-play solution to high-dimensional MSk control. Instead, three generalizable design principles emerged from the successful solutions. \textbf{First}, \textit{task-specific structure} was important for reducing exploration and credit-assignment difficulties. To handle the highly dynamic nature of incoming table tennis balls, all winning solutions moved beyond purely end-to-end learning and incorporated explicit physics-based modelling and task-specific priors. 
\textbf{Second}, \textit{hierarchical decomposition} helped make complex motor tasks more tractable. ActingAI, BioSyn, and LNSGroup separated trajectory planning from muscle-level execution, while LNSGroup further demonstrated the advantages of specialized forehand and backhand controllers over a single universal policy. Similarly, Servette MyoClub progressively built upon foundational locomotion skills~\cite{simos2025kinesis} before acquiring task-specific kicking behaviors. These approaches share structural similarities with biological motor learning, where foundational motor capabilities and high-level movement strategies support the execution of complex motor skills. More broadly, the results suggest that separating task-level planning from motor execution, or reusing previously acquired skills, can reduce the difficulty of learning complex muscle-driven behaviors.
\textbf{Lastly,} \textit{observation and action representations} influenced both learning efficiency and control expressiveness. ActingAI demonstrated the importance of including muscle activation states in policy observations, while BioSyn showed that removing planned striking targets substantially increased exploration difficulty. To manage the high-dimensional muscle action space, ActingAI utilized state-space compression, while BioSyn employed biologically inspired muscle-synergy subnetworks~\cite{Bizzi2013,Guimares2020,chiappa2023latent}. In contrast, LNSGroup employed CrossQ-based reinforcement learning for muscle-level stroke control without introducing an explicit learned low-dimensional muscle-synergy representation~\cite{bhattcrossq, wei2026scalable}, preserving control expressiveness but resulting in higher \textit{Effort} scores (Table~\ref{table:results}). In the Soccer track, Servette MyoClub leveraged imitation learning~\cite{simos2025kinesis} to start with a controller that already encodes a plausible motion prior, and thus did not require any explicit constraining of the action space. A post-hoc analysis of the functional dimensionality further showed that even synergies that explain only a small amount of variance are important for successful control (Appendix~\ref{sec:servette-dim-analysis}), which extends recent modeling results from dexterous hand control~\cite{chiappa2024acquiring,chiappa2025arnold0}. Overall, these insights suggest that informative observations are essential, whereas action-space reduction introduces a trade-off between learning efficiency and the expressiveness of muscle control, in part because it may discard low-variance components that still matter for control.

\paragraph{Unsuccessful Strategies}
The winning teams also reported several unsuccessful approaches. For example, LNSGroup found that successful approaches in previous challenges such as DynSyn~\cite{Kaibo_2024}, as well as CrossQ~\cite{cross-Q} without sufficient task-specific guidance, could not directly solve the Table Tennis task. Sparse rewards provided insufficient credit assignment, while dense tracking objectives could produce contact without valid returns. Their attempts to train a universal forehand--backhand controller also remained unreliable despite additional training and curriculum learning, emphasizing the advantages of policy specialization. Similarly, Servette MyoClub found that direct PPO training from scratch failed to establish stable locomotion, achieving a 0\% scoring rate. Those unsuccessful attempts showed that established principles such as muscle synergies, imitation learning, and reward shaping remained useful, but required adaptation to the more dynamic tasks introduced in 2025. 

\paragraph{Competition-Wide Analysis and Limitations}
Beyond the winning teams, the broader competition results reveal the limitations of current musculoskeletal control methods. In Table Tennis Phase~1, the 11 teams scoring above baseline averaged $76.3 \pm 24.9$ (median: $90$), whereas the nine teams achieving non-zero scores in Phase~2 averaged only $30.6 \pm 26.6$, with four scoring above $50$. Several strong Phase~1 teams deteriorated substantially under randomized incoming-ball conditions; for example, CIAO's score dropped from $96\%$ to $1\%$. These results demonstrate that strong performance in simplified task settings often fails to generalize to more complex and dynamic conditions. The Soccer track was even more challenging, with only Servette MyoClub achieving a non-zero scoring rate. Nevertheless, its resulting behavior shows a slow approach towards the ball and produces a small kick, rather than demonstrating diverse athletic movements. This performance gap reflects the distinction between optimizing a competition objective and reproducing the biomechanical characteristics of human performance, including coordinated kinetic-chain sequencing and muscle--tendon stretch-shortening dynamics~\cite{Jelusic1992,Lees2010}. While the winning solutions establish the feasibility of muscle-driven sports control, robust generalization and diverse, human-like motor behavior remain open challenges. 

\paragraph{Computational Scalability and Future Directions}
Computational cost represents an additional barrier to scalable musculoskeletal learning. Despite both achieving a $100\%$ success rate in Phase 1, Servette MyoClub required approximately 14 days to train its Phase~1 Soccer policy, while the Acting AI's Table Tennis policy required 51 hours, both using the CPU-based MyoSuite environment. As the community scales toward high-dimensional, full-body musculoskeletal control, this CPU bottleneck becomes prohibitive. Recent GPU-accelerated frameworks~\cite{Li2026MuscleMimic,Wei2026} offer opportunities to increase simulation throughput and enable more extensive experimentation with algorithms, reward formulations, and training curricula. Although these developments were not part of the official 2025 challenge, this paradigm shift paves the way for a fully GPU-accelerated and data-driven era in biomechanical simulation and future editions of MyoChallenge.

\section{Conclusion and Future Challenges}

Digital twins of humans are essential tools for understanding neuromotor control, muscle recruitment patterns, athletic performance, and rehabilitation strategies. In this paper, we present MyoChallenge 2025, a competition benchmarking the gap between human athletic abilities and current MSk simulation capabilities. The winning solutions demonstrated diverse approaches, including physics-based priors, muscle synergies, expert distillation, and behavior cloning. Despite these advances, scalable full-body control remains limited by both computational and algorithmic challenges in this edition. Future challenges will move toward GPU-based simulation to reduce computational bottlenecks, while continuing to lower barriers to participation, especially for underrepresented groups, and encourage the global community to advance neuromotor control and human–machine interaction.


\begin{ack}
We would like to acknowledge support for this competition from the University of Twente Techmed and DSI, Google Cloud Computing, MyoLab, Delsys, Ocon - Orthopesische Kliniek, Ocon -- Sportmedische Kliniek, and EU ERC StG Interact. Special thanks goes to Dhruv Batra, Ram Racmrakhya, Deshraj Yadav, and Rishabh Jain for help with the EvalAI platform. 
A.M.: Swiss SNF grant (310030$\_$212516) and Simons foundation (SFI-AN-NC-SCN-00007276-14).
G.D. and C.W.: NSERC Discovery Grant and NFRF: Exploration. B. K. H.: Imperial-Meta Wearable Neural Interfaces Research Centre.

The authors declare no competing financial interests or relationships that could be perceived as influencing the work presented in this paper.
\end{ack}

\renewcommand*{\bibfont}{\small}
\bibliographystyle{unsrt}
\bibliography{References}

@misc{omniH2O,
  doi = {10.48550/ARXIV.2406.08858},
  url = {https://arxiv.org/abs/2406.08858},
  author = {He,  Tairan and Luo,  Zhengyi and He,  Xialin and Xiao,  Wenli and Zhang,  Chong and Zhang,  Weinan and Kitani,  Kris and Liu,  Changliu and Shi,  Guanya},
  title = {OmniH2O: Universal and Dexterous Human-to-Humanoid Whole-Body Teleoperation and Learning},
  publisher = {arXiv},
  year = {2024},
  copyright = {arXiv.org perpetual,  non-exclusive license}
}

@article{he2025asap,
          title={ASAP: Aligning Simulation and Real-World Physics for Learning Agile Humanoid Whole-Body Skills},
          author={He, Tairan and Gao, Jiawei and Xiao, Wenli and Zhang, Yuanhang and Wang, Zi and Wang, Jiashun and Luo, Zhengyi and He, Guanqi and Sobanbabu, Nikhil and Pan, Chaoyi and Yi, Zeji and Qu, Guannan and Kitani, Kris and Hodgins, Jessica and Fan, Linxi "Jim" and Zhu, Yuke and Liu, Changliu and Shi, Guanya},
          journal={arXiv preprint arXiv:2502.01143},
          year={2025}
        }

@misc{cross-Q,
  doi = {10.48550/ARXIV.1902.05605},
  url = {https://arxiv.org/abs/1902.05605},
  author = {Bhatt,  Aditya and Palenicek,  Daniel and Belousov,  Boris and Argus,  Max and Amiranashvili,  Artemij and Brox,  Thomas and Peters,  Jan},
  title = {CrossQ: Batch Normalization in Deep Reinforcement Learning for Greater Sample Efficiency and Simplicity},
  publisher = {arXiv},
  year = {2019},
  copyright = {arXiv.org perpetual,  non-exclusive license}
}

@article{Christophy2011,
  title = {A Musculoskeletal model for the lumbar spine},
  volume = {11},
  ISSN = {1617-7940},
  url = {http://dx.doi.org/10.1007/s10237-011-0290-6},
  DOI = {10.1007/s10237-011-0290-6},
  number = {1–2},
  journal = {Biomechanics and Modeling in Mechanobiology},
  publisher = {Springer Science and Business Media LLC},
  author = {Christophy,  Miguel and Faruk Senan,  Nur Adila and Lotz,  Jeffrey C. and O’Reilly,  Oliver M.},
  year = {2011},
  month = feb,
  pages = {19–34}
}

@article{su2025hitter,
  title={Hitter: A humanoid table tennis robot via hierarchical planning and learning},
  author={Su, Zhi and Zhang, Bike and Rahmanian, Nima and Gao, Yuman and Liao, Qiayuan and Regan, Caitlin and Sreenath, Koushil and Sastry, S Shankar},
  journal={arXiv preprint arXiv:2508.21043},
  year={2025}
}

@article{liu2026biosyngrasp,
  author={Liu, Siyuan and Jiang, Bo and Han, Lijun and Zhong, Shanlin and Song, Ci and Liu, Huaping and Chen, Jiahao},
  journal={IEEE Transactions on Automation Science and Engineering}, 
  title={BioSynGrasp: Bio-Inspired Structured Reinforcement Learning With Synergetic Exploration and Human Demonstration for Dexterous Grasping of Musculoskeletal Robot}, 
  year={2026},
  volume={23}, 
  pages={6102-6116} 
}

@article{zhao2026diff,
  title={Diff-Muscle: Efficient Learning for Musculoskeletal Robotic Table Tennis},
  author={Zhao, Wentao and Guo, Jun and Huang, Kangyao and Liu, Xin and Liu, Huaping},
  journal={arXiv preprint arXiv:2603.08617},
  year={2026}
}

@book{sira2018differentially,
  title={Differentially flat systems},
  author={Sira-Ramirez, Hebertt and Agrawal, Sunil K},
  year={2018},
  publisher={Crc Press}
}

@article{sartori_neural_2016,
	title = {Neural Data-Driven Musculoskeletal Modeling for Personalized Neurorehabilitation Technologies},
	volume = {63},
	issn = {1558-2531},
	doi = {10.1109/TBME.2016.2538296},
	pages = {879--893},
	number = {5},
    year = {2016},
	journaltitle = {{IEEE} transactions on bio-medical engineering},
	journal = {{IEEE} transactions on bio-medical engineering},
	shortjournal = {{IEEE} Trans Biomed Eng},
	author = {Sartori, Massimo and Llyod, David G. and Farina, Dario},
	date = {2016-05},
	pmid = {27046865},
}

@article{
Ramdya_Ijspeert_2023,
author = {Pavan Ramdya  and Auke Jan Ijspeert },
title = {The neuromechanics of animal locomotion: From biology to robotics and back},
journal = {Science Robotics},
volume = {8},
number = {78},
pages = {eadg0279},
year = {2023},
doi = {10.1126/scirobotics.adg0279},
URL = {https://www.science.org/doi/abs/10.1126/scirobotics.adg0279},
eprint = {https://www.science.org/doi/pdf/10.1126/scirobotics.adg0279}}

@ARTICLE{Sartori_2023,
  author={Sartori, Massimo},
  journal={IEEE Robotics \& Automation Magazine}, 
  title={Advancing Wearable Robotics for Shaping the Human Musculoskeletal System [Young Professionals]}, 
  year={2023},
  volume={30},
  number={3},
  pages={164-165},
  doi={10.1109/MRA.2023.3293338}}

@book{Valero_Cuevas_2024,
	author={ Francisco J. Valero-Cuevas},
    title = {Fundamentals of Neuromechanics },
    year = {2024},
    publsiher = {Springer},
	book = { Biosystems & Biorobotics},
    url = {https://link.springer.com/book/10.1007/978-1-4471-6747-1},
	urldate = {2024-03-27},
}

@inproceedings{
wang2025myochallenge,
title={MyoChallenge 2024: A New Benchmark for Physiological Dexterity and Agility in Bionic Humans},
author={Cheryl Wang and Chun Kwang Tan and Balint K Hodossy and Shirui Lyu and Pierre Schumacher and James Heald and Kai Biegun and Samo Hromadka and Maneesh Sahani and Gunwoo Park and Beomsoo Shin and JongHyun Park and SEUNGBUM KOO and Chenhui Zuo and Chengtian Ma and Yanan Sui and Nicklas Hansen and Stone Tao and Yuan Gao and Hao Su and Seungmoon Song and Letizia Gionfrida and Massimo Sartori and Guillaume Durandau and Vikash Kumar and Vittorio Caggiano},
booktitle={The Thirty-ninth Annual Conference on Neural Information Processing Systems Datasets and Benchmarks Track},
year={2025},
url={https://openreview.net/forum?id=1dSLbhErNv}
}

@misc{www_mobl,
    author = {SimTK},
  title = {Upper Extremity Dynamic Model},
  howpublished = {\url{https://simtk.org/projects/upexdyn}},
  note = {}
}

@Misc{MyoChallenge2023,
 author =	{Caggiano, Vittorio AND Wang, Huawei AND Durandau, Guillaume AND Song, Seungmoon AND Chun Kwang, Tan AND Cameron, Berg AND Pierre, Schumacher AND Sartori, Massimo AND Kumar, Vikash},
 title =	{MyoChallenge 2023: Towards Human-Level Dexterity and Agility},
 howpublished = {\url{https://sites.google.com/view/myosuite/myochallenge/myochallenge-2023}},
 year = 	{2023}
}

@inproceedings{myosim,
  title={MyoSim: Fast and physiologically realistic MuJoCo models for musculoskeletal and exoskeletal studies},
  author={Wang, Huawei and Caggiano, Vittorio and Durandau, Guillaume and Sartori, Massimo and Kumar, Vikash},
  booktitle={2022 International Conference on Robotics and Automation (ICRA)},
  pages={8104--8111},
  year={2022},
  organization={IEEE}
}

@article{chiappa2024acquiring,
  title={Acquiring musculoskeletal skills with curriculum-based reinforcement learning},
  author={Chiappa, Alberto Silvio and Tano, Pablo and Patel, Nisheet and Ingster, Abigail and Pouget, Alexandre and Mathis, Alexander},
  journal={bioRxiv},
  pages={2024--01},
  year={2024},
  publisher={Cold Spring Harbor Laboratory}
}

@inproceedings{
    IsaacGym_2021,
    title={Isaac Gym: High Performance {GPU} Based Physics Simulation For Robot Learning},
    author={Viktor Makoviychuk and Lukasz Wawrzyniak and Yunrong Guo and Michelle Lu and Kier Storey and Miles Macklin and David Hoeller and Nikita Rudin and Arthur Allshire and Ankur Handa and Gavriel State},
    booktitle={Thirty-fifth Conference on Neural Information Processing Systems Datasets and Benchmarks Track (Round 2)},
    year={2021},
    url={https://openreview.net/forum?id=fgFBtYgJQX_}
}

@article{song2021deep,
  title={Deep reinforcement learning for modeling human locomotion control in neuromechanical simulation},
  author={Song, Seungmoon and Kidzi{\'n}ski, {\L}ukasz and Peng, Xue Bin and Ong, Carmichael and Hicks, Jennifer and Levine, Sergey and Atkeson, Christopher G and Delp, Scott L},
  journal={Journal of neuroengineering and rehabilitation},
  volume={18},
  number={1},
  pages={1--17},
  year={2021},
  publisher={Springer}
}

@inproceedings{Erez_2015,
  author    = {Tom Erez and
               Yuval Tassa and
               Emanuel Todorov},
  title     = {Simulation tools for model-based robotics: Comparison of Bullet, Havok,
               MuJoCo, {ODE} and PhysX},
  booktitle = {{IEEE} International Conference on Robotics and Automation, {ICRA}
               2015},
  pages     = {4397--4404},
  publisher = {{IEEE}},
  year      = {2015},
  url       = {https://doi.org/10.1109/ICRA.2015.7139807},
  doi       = {10.1109/ICRA.2015.7139807},
  bibsource = {dblp computer science bibliography, https://dblp.org}
}

@inproceedings{ikkala2020,
  title={Converting biomechanical models from opensim to Mujoco},
  author={Ikkala, Aleksi and H{\"a}m{\"a}l{\"a}inen, Perttu},
  booktitle={Converging Clinical and Engineering Research on Neurorehabilitation IV: Proceedings of the 5th International Conference on Neurorehabilitation (ICNR2020), October 13--16, 2020},
  pages={277--281},
  year={2022},
  organization={Springer}
}

@inproceedings{myosuite,
  title={MyoSuite: A contact-rich simulation suite for musculoskeletal motor control},
  author={Caggiano, Vittorio and Wang, Huawei and Durandau, Guillaume and Sartori, Massimo, Kumar and Vikash},
  booktitle={4th Annual Conference on Learning for Dynamics and Control (L4DC)},
  year={2022},
  organization={Proceedings of Machine Learning Research }
}

@misc{caggiano_myosuite_2022,
  doi = {10.48550/ARXIV.2205.13600},
  
  url = {https://arxiv.org/abs/2205.13600},
  
  author = {Caggiano, Vittorio and Wang, Huawei and Durandau, Guillaume and Sartori, Massimo and Kumar, Vikash},
  
  title = {MyoSuite -- A contact-rich simulation suite for musculoskeletal motor control},
  
  publisher = {arXiv},
  
  year = {2022},
  
  copyright = {Creative Commons Attribution Non Commercial No Derivatives 4.0 International}
}

@article{saul2015benchmarking,
  title={Benchmarking of dynamic simulation predictions in two software platforms using an upper limb musculoskeletal model},
  author={Saul, Katherine R and Hu, Xiao and Goehler, Craig M and Vidt, Meghan E and Daly, Melissa and Velisar, Anca and Murray, Wendy M},
  journal={Computer methods in biomechanics and biomedical engineering},
  volume={18},
  number={13},
  pages={1445--1458},
  year={2015},
  publisher={Taylor \& Francis}
}

@article{song2015neural,
  title={A neural circuitry that emphasizes spinal feedback generates diverse behaviours of human locomotion},
  author={Song, Seungmoon and Geyer, Hartmut},
  journal={The Journal of physiology},
  volume={593},
  number={16},
  pages={3493--3511},
  year={2015},
  publisher={Wiley Online Library}
}

@inproceedings{todorov2012mujoco,
  title={Mujoco: A physics engine for model-based control},
  author={Todorov, Emanuel and Erez, Tom and Tassa, Yuval},
  booktitle={2012 IEEE/RSJ International Conference on Intelligent Robots and Systems},
  pages={5026--5033},
  year={2012},
  organization={IEEE}
}

@inproceedings{kidzinski2018learning,
  title={Learning to run challenge: Synthesizing physiologically accurate motion using deep reinforcement learning},
  author={Kidzi{\'n}ski, {\L}ukasz and Mohanty, Sharada P and Ong, Carmichael F and Hicks, Jennifer L and Carroll, Sean F and Levine, Sergey and Salath{\'e}, Marcel and Delp, Scott L},
  booktitle={The NIPS'17 Competition: Building Intelligent Systems},
  pages={101--120},
  year={2018},
  organization={Springer}
}

@ARTICLE{Rajagopal2016,
  author={Rajagopal, Apoorva and Dembia, Christopher L. and DeMers, Matthew S. and Delp, Denny D. and Hicks, Jennifer L. and Delp, Scott L.},
  journal={IEEE Transactions on Biomedical Engineering}, 
  title={Full-Body Musculoskeletal Model for Muscle-Driven Simulation of Human Gait}, 
  year={2016},
  volume={63},
  number={10},
  pages={2068-2079},
  doi={10.1109/TBME.2016.2586891}}

@inproceedings{caggiano2024myochallenge,
  title={MyoChallenge 2023: Towards Human-Level Dexterity and Agility},
  author={Caggiano, Vittorio and Durandau, Guillaume and Wang, Huiyi and Tan, Chun Kwang and Schumacher, Pierre and Wang, Huawei and Chiappa, Alberto Silvio and Marin Vargas, Alessandro and Mathis, Alexander and Won, Jungdam and Park, Jungnam and Park, Gunwoo and Shin, Beomsoo and Kim, Minseung and Koo, Seungbum and Yang, Zhuo and Dang, Wei and Cai, Heng and Song, Jianfei and Song, Seungmoon},
  booktitle={NeurIPS 2024 Track Datasets and Benchmarks},
  year={2024},
  url={https://openreview.net/forum?id=3A84lx1JFh}
}

@inproceedings{Walia2025,
  title = {Myoback: A Musculoskeletal Model of the Human Back with Integrated Exoskeleton},
  url = {http://dx.doi.org/10.1109/ICORR66766.2025.11063132},
  DOI = {10.1109/icorr66766.2025.11063132},
  booktitle = {2025 International Conference On Rehabilitation Robotics (ICORR)},
  publisher = {IEEE},
  author = {Walia,  Rohan and Billot,  Morgane and Garzon-Aguirre,  Kevin and Subramanian,  Swathika and Wang,  Huiyi and Refai,  Mohamed Irfan and Durandau,  Guillaume},
  year = {2025},
  month = may,
  pages = {128–135}
}

@article{won2022physics,
author = {Won, Jungdam and Gopinath, Deepak and Hodgins, Jessica},
title = {Physics-based character controllers using conditional VAEs},
year = {2022},
issue_date = {July 2022},
publisher = {Association for Computing Machinery},
address = {New York, NY, USA},
volume = {41},
number = {4},
issn = {0730-0301},
url = {https://doi.org/10.1145/3528223.3530067},
doi = {10.1145/3528223.3530067},
journal = {ACM Trans. Graph.},
month = jul,
articleno = {96},
numpages = {12},
}

@article{Azocar2020,
  title = {Design and clinical implementation of an open-source bionic leg},
  volume = {4},
  ISSN = {2157-846X},
  url = {http://dx.doi.org/10.1038/s41551-020-00619-3},
  DOI = {10.1038/s41551-020-00619-3},
  number = {10},
  journal = {Nature Biomedical Engineering},
  publisher = {Springer Science and Business Media LLC},
  author = {Azocar,  Alejandro F. and Mooney,  Luke M. and Duval,  Jean-Fran\c{c}ois and Simon,  Ann M. and Hargrove,  Levi J. and Rouse,  Elliott J.},
  year = {2020},
  month = oct,
  pages = {941–953}
}

@article{Yu2021,
  title = {Clinical evaluation of the revolutionizing prosthetics modular prosthetic limb system for upper extremity amputees},
  volume = {11},
  ISSN = {2045-2322},
  url = {http://dx.doi.org/10.1038/s41598-020-79581-8},
  DOI = {10.1038/s41598-020-79581-8},
  number = {1},
  journal = {Scientific Reports},
  publisher = {Springer Science and Business Media LLC},
  author = {Yu,  Kristin E. and Perry,  Briana N. and Moran,  Courtney W. and Armiger,  Robert S. and Johannes,  Matthew S. and Hawkins,  Abigail and Stentz,  Lauren and Vandersea,  Jamie and Tsao,  Jack W. and Pasquina,  Paul F.},
  year = {2021},
  month = jan 
}

@article{Song2020,
  title = {Deep reinforcement learning for modeling human locomotion control in neuromechanical simulation},
  url = {http://dx.doi.org/10.1101/2020.08.11.246801},
  DOI = {10.1101/2020.08.11.246801},
  publisher = {Cold Spring Harbor Laboratory},
  journal= {Journal of neuroengineering and rehabilitation},
  author = {Song,  Seungmoon and Kidziński,  Łukasz and Peng,  Xue Bin and Ong,  Carmichael and Hicks,  Jennifer and Levine,  Sergey and Atkeson,  Christopher G. and Delp,  Scott L.},
  year = {2020},
  month = aug 
}

@misc{Kidzinski2019,
  doi = {10.48550/ARXIV.1902.02441},
  url = {https://arxiv.org/abs/1902.02441},
  author = {Kidziński,  Łukasz and Ong,  Carmichael and Mohanty,  Sharada Prasanna and Hicks,  Jennifer and Carroll,  Sean F. and Zhou,  Bo and Zeng,  Hongsheng and Wang,  Fan and Lian,  Rongzhong and Tian,  Hao and Jaśkowski,  Wojciech and Andersen,  Garrett and Lykkebø,  Odd Rune and Toklu,  Nihat Engin and Shyam,  Pranav and Srivastava,  Rupesh Kumar and Kolesnikov,  Sergey and Hrinchuk,  Oleksii and Pechenko,  Anton and Ljungstr\"{o}m,  Mattias and Wang,  Zhen and Hu,  Xu and Hu,  Zehong and Qiu,  Minghui and Huang,  Jun and Shpilman,  Aleksei and Sosin,  Ivan and Svidchenko,  Oleg and Malysheva,  Aleksandra and Kudenko,  Daniel and Rane,  Lance and Bhatt,  Aditya and Wang,  Zhengfei and Qi,  Penghui and Yu,  Zeyang and Peng,  Peng and Yuan,  Quan and Li,  Wenxin and Tian,  Yunsheng and Yang,  Ruihan and Ma,  Pingchuan and Khadka,  Shauharda and Majumdar,  Somdeb and Dwiel,  Zach and Liu,  Yinyin and Tumer,  Evren and Watson,  Jeremy and Salathé,  Marcel and Levine,  Sergey and Delp,  Scott},
  title = {Artificial Intelligence for Prosthetics - challenge solutions},
  publisher = {arXiv},
  year = {2019},
  copyright = {arXiv.org perpetual,  non-exclusive license}
}

@inproceedings{ikkala2022breathing,
  title={Breathing Life Into Biomechanical User Models},
  author={Ikkala, Aleksi and Fischer, Florian and Klar, Markus and Bachinski, Miroslav and Fleig, Arthur and Howes, Andrew and H{\"a}m{\"a}l{\"a}inen, Perttu and M{\"u}ller, J{\"o}rg and Murray-Smith, Roderick and Oulasvirta, Antti},
  booktitle={Proceedings of the 35th Annual ACM Symposium on User Interface Software and Technology},
  pages={1--14},
  year={2022}
}

@inproceedings{park2022generative,
  title={Generative gaitnet},
  author={Park, Jungnam and Min, Sehee and Chang, Phil Sik and Lee, Jaedong and Park, Moon Seok and Lee, Jehee},
  booktitle={ACM SIGGRAPH 2022 Conference Proceedings},
  pages={1--9},
  year={2022}
}

@inproceedings{schumacher2023:deprl,
  title = {DEP-RL: Embodied Exploration for Reinforcement Learning in Overactuated and Musculoskeletal Systems},
  author = {Schumacher, Pierre and Haeufle, Daniel F.B. and B{\"u}chler, Dieter and Schmitt, Syn and Martius, Georg},
  booktitle = {Proceedings of the Eleventh International Conference on Learning Representations (ICLR)},
  month = may,
  year = {2023},
  doi = {},
  url = {https://openreview.net/forum?id=C-xa_D3oTj6},
  month_numeric = {5}
}

@Misc{MyoChallenge2022,
 author =	{Caggiano, Vittorio AND Wang, Huawei AND Durandau, Guillaume AND Song, Seungmoon AND Tassa, Yuval AND   Sartori, Massimo AND Kumar, Vikash},
 title =	{MyoChallenge: Learning contact-rich manipulation using a musculoskeletal hand},
 howpublished = {\url{ https://sites.google.com/view/myochallenge }},
 year = 	{2022}
}

@misc{Geijtenbeek2021Hyfydy,
  author = {Geijtenbeek, Thomas},
  title = {The {Hyfydy} Simulation Software},
  year = {2021},
  month = {11},
  url = {https://hyfydy.com},
  note = {\url{https://hyfydy.com}}
}

@article{opensim,
  author = {Delp, Scott L. and Anderson, Frank C. and Arnold, Allison S. and Loan, Peter and Habib, Ayman and John, Chand T. and Guendelman, Eran and Thelen, Darryl G.},
  ee = {https://www.wikidata.org/entity/Q33306019},
  journal = {IEEE Trans. Biomed. Eng.},
  number = 11,
  pages = {1940-1950},
  title = {OpenSim: Open-Source Software to Create and Analyze Dynamic Simulations of Movement.},
  url = {http://dblp.uni-trier.de/db/journals/tbe/tbe54.html#DelpAALHJGT07},
  volume = 54,
  year = 2007
}

@misc{zhuang2023,
      title={Robot Parkour Learning}, 
      author={Ziwen Zhuang and Zipeng Fu and Jianren Wang and Christopher Atkeson and Soeren Schwertfeger and Chelsea Finn and Hang Zhao},
      year={2023},
      eprint={2309.05665},
      archivePrefix={arXiv},
      primaryClass={cs.RO},
      url={https://arxiv.org/abs/2309.05665}, 
}

@inproceedings{Truong_2024, series={SA ’24},
   title={PDP: Physics-Based Character Animation via Diffusion Policy},
   url={http://dx.doi.org/10.1145/3680528.3687683},
   DOI={10.1145/3680528.3687683},
   booktitle={SIGGRAPH Asia 2024 Conference Papers},
   publisher={ACM},
   author={Truong, Takara Everest and Piseno, Michael and Xie, Zhaoming and Liu, Karen},
   year={2024},
   month=dec, pages={1–10},
   collection={SA ’24} }

@misc{zhang2025generative,
    title         = {A Generative System for Robot-to-Human Handovers: from Intent Inference to Spatial Configuration Imagery}, 
    author        = {Hanxin Zhang and Abdulqader Dhafer and Zhou Daniel Hao and Hongbiao Dong},
    year          = {2025},
    eprint        = {2503.03579},
    archivePrefix = {arXiv},
    primaryClass  = {cs.RO},
    url           = {https://arxiv.org/abs/2503.03579}, 
}

@article{Bizzi2013,
  title = {The neural origin of muscle synergies},
  volume = {7},
  ISSN = {1662-5188},
  url = {http://dx.doi.org/10.3389/fncom.2013.00051},
  DOI = {10.3389/fncom.2013.00051},
  journal = {Frontiers in Computational Neuroscience},
  publisher = {Frontiers Media SA},
  author = {Bizzi,  Emilio and Cheung,  Vincent C. K.},
  year = {2013}
}

@article{Jelusic1992,
author = {Jelusic, V. and Jaric, Slobodan and Kukolj, M.},
year = {1992},
month = {01},
pages = {231-238},
title = {Effects of the stretch-shortening strength training on kicking performance in soccer players},
volume = {22},
journal = {J Hum Mov Stud}
}

@article{Lees2010,
  title = {The biomechanics of kicking in soccer: A review},
  volume = {28},
  ISSN = {1466-447X},
  url = {http://dx.doi.org/10.1080/02640414.2010.481305},
  DOI = {10.1080/02640414.2010.481305},
  number = {8},
  journal = {Journal of Sports Sciences},
  publisher = {Informa UK Limited},
  author = {Lees,  A. and Asai,  T. and Andersen,  T. B. and Nunome,  H. and Sterzing,  T.},
  year = {2010},
  month = jun,
  pages = {805–817}
}

@article{Li2026MuscleMimic,
  title={Towards Embodied AI with MuscleMimic: Unlocking full-body musculoskeletal motor learning at scale},
  author={Li, Chengkun and Wang, Cheryl and Ziliotto, Bianca and Simos, Merkourios and Kovecses, Jozsef and Durandau, Guillaume and Mathis, Alexander},
  journal={arXiv preprint arXiv:2603.25544},
  year={2026}
}

@article{Guimares2020,
  title = {Freezing Degrees of Freedom During Motor Learning: A Systematic Review},
  volume = {24},
  ISSN = {1543-2696},
  url = {http://dx.doi.org/10.1123/mc.2019-0060},
  DOI = {10.1123/mc.2019-0060},
  number = {3},
  journal = {Motor Control},
  publisher = {Human Kinetics},
  author = {Guimarães,  Anderson Nascimento and Ugrinowitsch,  Herbert and Dascal,  Juliana Bayeux and Porto,  Alessandra Beggiato and Okazaki,  Victor Hugo Alves},
  year = {2020},
  month = jul,
  pages = {457–471}
}

@article{Wei2026,
  title={Scaling Whole-Body Human Musculoskeletal Behavior Emulation for Specificity and Diversity},
  author={Wei, Yunyue and Zuo, Chenhui and Zhuang, Shanning and Gong, Haixin and Liu, Yaming and Sui, Yanan},
  journal={arXiv preprint arXiv:2603.29332},
  year={2026}
}

@article{liu2024realdex,
  title={Realdex: Towards human-like grasping for robotic dexterous hand},
  author={Liu, Yumeng and Yang, Yaxun and Wang, Youzhuo and Wu, Xiaofei and Wang, Jiamin and Yao, Yichen and Schwertfeger, S{\"o}ren and Yang, Sibei and Wang, Wenping and Yu, Jingyi and others},
  journal={arXiv preprint arXiv:2402.13853},
  year={2024}
}

@article{
Liu_scirobotics_2022,
author = {Siqi Liu  and Guy Lever  and Zhe Wang  and Josh Merel  and S. M. Ali Eslami  and Daniel Hennes  and Wojciech M. Czarnecki  and Yuval Tassa  and Shayegan Omidshafiei  and Abbas Abdolmaleki  and Noah Y. Siegel  and Leonard Hasenclever  and Luke Marris  and Saran Tunyasuvunakool  and H. Francis Song  and Markus Wulfmeier  and Paul Muller  and Tuomas Haarnoja  and Brendan Tracey  and Karl Tuyls  and Thore Graepel  and Nicolas Heess },
title = {From motor control to team play in simulated humanoid football},
journal = {Science Robotics},
volume = {7},
number = {69},
pages = {eabo0235},
year = {2022},
doi = {10.1126/scirobotics.abo0235},
URL = {https://www.science.org/doi/abs/10.1126/scirobotics.abo0235},
eprint = {https://www.science.org/doi/pdf/10.1126/scirobotics.abo0235},
}

@article{chiappa2023latent,
  title={Latent Exploration for Reinforcement Learning},
          author={Chiappa, Alberto Silvio and Vargas, Alessandro Marin and Huang, Ann Zixiang and Mathis, Alexander},
  journal={Advances in Neural Information Processing Systems},
  year={2023}
}

@inproceedings{simos2025kinesis,
  title={Kinesis: Motion imitation for human musculoskeletal locomotion},
  author={Simos, Merkourios and Chiappa, Alberto Silvio and Mathis, Alexander},
  booktitle={2026 IEEE international conference on robotics and automation (ICRA)},
  year={2026},
  organization={IEEE}
}

@inproceedings{mandery2015kit,
  title={The KIT whole-body human motion database},
  author={Mandery, Christian and Terlemez, {\"O}mer and Do, Martin and Vahrenkamp, Nikolaus and Asfour, Tamim},
  booktitle={2015 International Conference on Advanced Robotics (ICAR)},
  pages={329--336},
  year={2015},
  organization={IEEE}
}

@inproceedings{ross2011reduction,
  title={A reduction of imitation learning and structured prediction to no-regret online learning},
  author={Ross, St{\'e}phane and Gordon, Geoffrey and Bagnell, Drew},
  booktitle={Proceedings of the fourteenth international conference on artificial intelligence and statistics},
  pages={627--635},
  year={2011},
  organization={JMLR Workshop and Conference Proceedings}
}

@article{chiappa2025arnold0,
  title   = {Arnold: a generalist muscle transformer policy},
  author  = {Alberto Silvio Chiappa and Boshi An and Merkourios Simos and Chengkun Li and Alexander Mathis},
  year    = {2025},
  journal = {arXiv preprint arXiv: 2508.18066}
}

@inproceedings{bhattcrossq,
  title={CrossQ: Batch Normalization in Deep Reinforcement Learning for Greater Sample Efficiency and Simplicity},
  year={2024},
  author={Bhatt, Aditya and Palenicek, Daniel and Belousov, Boris and Argus, Max and Amiranashvili, Artemij and Brox, Thomas and Peters, Jan},
  booktitle={The Twelfth International Conference on Learning Representations}
}

@book{williams2006gaussian,
  title={Gaussian processes for machine learning},
  author={Williams, Christopher KI and Rasmussen, Carl Edward},
  volume={2},
  year={2006},
  publisher={MIT press Cambridge, MA}
}

@conference{AMASS:ICCV:2019,
  title = {{AMASS}: Archive of Motion Capture as Surface Shapes},
  author = {Mahmood, Naureen and Ghorbani, Nima and Troje, Nikolaus F. and Pons-Moll, Gerard and Black, Michael J.},
  booktitle = {International Conference on Computer Vision},
  pages = {5442--5451},
  month = oct,
  year = {2019},
  month_numeric = {10}
}

@misc{bones_seed_2026,
  author = {{Bones Studio}},
  title = {BONES-SEED: Skeletal Everyday Embodiment Dataset},
  year = {2026},
  publisher = {Hugging Face},
  howpublished = {\url{https://huggingface.co/datasets/bones-studio/seed}}
}

@article{Clancy2023,
  title = {Muscle-driven simulations and experimental data of cycling},
  volume = {13},
  ISSN = {2045-2322},
  url = {http://dx.doi.org/10.1038/s41598-023-47945-5},
  DOI = {10.1038/s41598-023-47945-5},
  number = {1},
  journal = {Scientific Reports},
  publisher = {Springer Science and Business Media LLC},
  author = {Clancy,  Caitlin E. and Gatti,  Anthony A. and Ong,  Carmichael F. and Maly,  Monica R. and Delp,  Scott L.},
  year = {2023},
  month = dec 
}

@article{Buffi2014,
  title = {Computing Muscle,  Ligament,  and Osseous Contributions to the Elbow Varus Moment During Baseball Pitching},
  volume = {43},
  ISSN = {1573-9686},
  url = {http://dx.doi.org/10.1007/s10439-014-1144-z},
  DOI = {10.1007/s10439-014-1144-z},
  number = {2},
  journal = {Annals of Biomedical Engineering},
  publisher = {Springer Science and Business Media LLC},
  author = {Buffi,  James H. and Werner,  Katie and Kepple,  Tom and Murray,  Wendy M.},
  year = {2014},
  month = oct,
  pages = {404–415}
}

@article{Attias2026,
  title = {Musculoskeletal modelling and predictive simulation of baseball pitching to improve performance and mitigate injury using forward dynamics and optimal control},
  ISSN = {1573-272X},
  url = {http://dx.doi.org/10.1007/s11044-026-10143-y},
  DOI = {10.1007/s11044-026-10143-y},
  journal = {Multibody System Dynamics},
  publisher = {Springer Science and Business Media LLC},
  author = {Attias,  Cedric E. and Uchida,  Thomas K. and Inkol,  Keaton and McPhee,  John},
  year = {2026},
  month = jan 
}

@techreport{myoskeleton,
  author      = {Vittorio Caggiano AND Vittorio La Barbera AND Andrea Prestia AND Ouassim Aouattah AND Pierre Schumacher AND Varun Joshi AND Vikash Kumar},
  title       = {MyoSkeleton: A Universal Human Skeletal Model},
  institution = {MyoLab Inc.},
  year        = {2024},
  type        = {White Paper},
  note        = {Available at: \url{https://github.com/myolab/myo_model}},
}

@article{Martin2013,
  title = {Upper limb joint kinetic analysis during tennis serve: Assessment of competitive level on efficiency and injury risks},
  volume = {24},
  ISSN = {1600-0838},
  url = {http://dx.doi.org/10.1111/sms.12043},
  DOI = {10.1111/sms.12043},
  number = {4},
  journal = {Scandinavian Journal of Medicine \& Science in Sports},
  publisher = {Wiley},
  author = {Martin,  C. and Bideau,  B. and Ropars,  M. and Delamarche,  P. and Kulpa,  R.},
  year = {2013},
  month = jan,
  pages = {700–707}
}

@inproceedings{Kaibo_2024,
author = {He, Kaibo and Zuo, Chenhui and Ma, Chengtian and Sui, Yanan},
title = {DynSyn: dynamical synergistic representation for efficient learning and control in overactuated embodied systems},
year = {2024},
publisher = {JMLR.org},
booktitle = {Proceedings of the 41st International Conference on Machine Learning},
articleno = {727},
numpages = {18},
location = {Vienna, Austria},
series = {ICML'24}
}

@misc{bostondynamics2023,
  author = {Boston Dynamics},
  title = {Atlas},
  year = {2023},
  howpublished = {\url{https://bostondynamics.com/atlas}},
}

@article{Wang2025,
  title = {Reinforcement Learning Identifies Age-Related Balance Strategy Shifts},
  volume = {33},
  ISSN = {1558-0210},
  url = {http://dx.doi.org/10.1109/TNSRE.2025.3619868},
  DOI = {10.1109/tnsre.2025.3619868},
  journal = {IEEE Transactions on Neural Systems and Rehabilitation Engineering},
  publisher = {Institute of Electrical and Electronics Engineers (IEEE)},
  author = {Wang,  Huiyi and Kovecses,  Jozsef and Durandau,  Guillaume},
  year = {2025},
  pages = {4078–4088}
}

@article{EvalAI2019,
    title   =  {EvalAI: Towards Better Evaluation Systems for AI Agents},
    author  =  {Deshraj Yadav and Rishabh Jain and Harsh Agrawal and Prithvijit
                Chattopadhyay and Taranjeet Singh and Akash Jain and Shiv Baran
                Singh and Stefan Lee and Dhruv Batra},
    journal = {arXiv},
    year    =  {2019},
    volume  =  {arXiv:1902.03570}
}

@inproceedings{wei2026scalable,
    title={Scalable Exploration for High-Dimensional Continuous Control via Value-Guided Flow},
    author={Wei, Yunyue and Zuo, Chenhui and Sui, Yanan},
    booktitle={The Fourteenth International Conference on Learning Representations},
    year={2026}
}

@book{bernstein1967coordination,
	title        = {The co-ordination and regulation of movements},
	author       = {Bernstein, Nikolai A.},
	year         = 1967,
	publisher    = {Oxford, New York, Pergamon Press},
	volume       = 1
}

@article{loeb2021learning,
  title={Learning to use muscles},
  author={Loeb, Gerald E},
  journal={Journal of human kinetics},
  volume={76},
  pages={9},
  year={2021}
}

@article{tresch2009case,
  title={The case for and against muscle synergies},
  author={Tresch, Matthew C and Jarc, Anthony},
  journal={Current opinion in neurobiology},
  volume={19},
  number={6},
  pages={601--607},
  year={2009},
  publisher={Elsevier}
}

\newpage
\section*{NeurIPS Paper Checklist}

\begin{enumerate}

\item {\bf Claims}
    \item[] Question: Do the main claims made in the abstract and introduction accurately reflect the paper's contributions and scope?
    \item[] Answer: \answerYes{} 
    \item[] Justification: The abstract and introduction rigorously align with the paper’s contributions: (1) a novel competition framework for benchmarking sports intelligence in MSk model, (2) detailed methodological analysis of competition results, and (3) quantitative improvements over prior iterations (e.g., task complexity, participant performance). (4) New state-of-the-art methods developed by the winning teams.

\item {\bf Limitations}
    \item[] Question: Does the paper discuss the limitations of the work performed by the authors?
    \item[] Answer: \answerYes{} 
    \item[] Justification: In Section \ref{sec:dis}, we analyze (1) the methodological limitations of each team’s approach and (2) the relationship between each team's methodological approach and their difference from true sports intelligence (3) the gaps between the current CPU-enabled environment and potential transition to GPU. We identify key challenges and propose actionable steps to bridge these gaps in future work.

\item {\bf Theory assumptions and proofs}
    \item[] Question: For each theoretical result, does the paper provide the full set of assumptions and a complete (and correct) proof?
    \item[] Answer: \answerNA{}.
    \item[] Justification: This paper does not include theoretical results.

    \item {\bf Experimental result reproducibility}
    \item[] Question: Does the paper fully disclose all the information needed to reproduce the main experimental results of the paper to the extent that it affects the main claims and/or conclusions of the paper (regardless of whether the code and data are provided or not)?
    \item[] Answer: \answerYes{}{} 
    \item[] Justification: Yes, we have included the methodology, performance, and setup of the baseline and each team's solutions in section \ref{sec:results}, Appendix \ref{app:tt}, \ref{app:actingai}, \ref{app:biosyn}, \ref{app:lnsgroup} and \ref{app:servette}.

\item {\bf Open access to data and code}
    \item[] Question: Does the paper provide open access to the data and code, with sufficient instructions to faithfully reproduce the main experimental results, as described in supplemental material?
    \item[] Answer: \answerYes{} 
    \item[] Justification: Yes, we have included the URL for code, instructions, baseline in the abstract, section \ref{app:tt} and Appendix \ref{appendix: tutorial&bl}. The solutions of the participants are provided base in their own section's Appendix \ref{app:actingai}, \ref{app:biosyn}, \ref{app:lnsgroup} and \ref{app:servette}.

\item {\bf Experimental setting/details}
    \item[] Question: Does the paper specify all the training and test details (e.g., data splits, hyperparameters, how they were chosen, type of optimizer) necessary to understand the results?
    \item[] Answer: \answerYes{} 
    \item[] Justification: We provide the necessary training setup and details in Appendix \ref{appendix:env} and Section \ref{section2},  as well as the individual's team set up in Appendix \ref{app:actingai}, \ref{app:biosyn}, \ref{app:lnsgroup} and \ref{app:servette}.

\item {\bf Experiment statistical significance}
    \item[] Question: Does the paper report error bars suitably and correctly defined or other appropriate information about the statistical significance of the experiments?
    \item[] Answer: \answerYes{}
    \item[] Justification: While this challenge doesn't report the statistical significance due to limited sample size, we use success rate and effort as the metric as detailed in Section \ref{sec:results}.

\item {\bf Experiments compute resources}
    \item[] Question: For each experiment, does the paper provide sufficient information on the computer resources (type of compute workers, memory, time of execution) needed to reproduce the experiments?
    \item[] Answer:  \answerYes{}
    \item[] Justification: The compute resources of the baseline and solutions are outlined within Appendix \ref{app:tt}, \ref{app:actingai}, \ref{app:biosyn}, \ref{app:lnsgroup} and \ref{app:servette}.
    
\item {\bf Code of ethics}
    \item[] Question: Does the research conducted in the paper conform, in every respect, with the NeurIPS Code of Ethics \url{https://neurips.cc/public/EthicsGuidelines}?
    \item[] Answer: \answerYes{} 
    \item[] Justification: The author confirmed that the paper follows the NeurIPS Code of Ethics.

\item {\bf Broader impacts}
    \item[] Question: Does the paper discuss both potential positive societal impacts and negative societal impacts of the work performed?
    \item[] Answer: \answerYes{} 
    \item[] Justification: Yes, our paper discusses how the benchmarking environment that we have provided can positively impact the community with neuroscience, biomechanics, and sports rehabilitation within the introduction and conclusion. We also discuss the gap between our current framework and solutions with true sports athletes. 
    
\item {\bf Safeguards}
    \item[] Question: Does the paper describe safeguards that have been put in place for responsible release of data or models that have a high risk for misuse (e.g., pre-trained language models, image generators, or scraped datasets)?
    \item[] Answer: \answerNA{}
    \item[] Justification: The paper poses no risk of data misuse.

\item {\bf Licenses for existing assets}
    \item[] Question: Are the creators or original owners of assets (e.g., code, data, models), used in the paper, properly credited and are the license and terms of use explicitly mentioned and properly respected?
    \item[] Answer: \answerYes{}
    \item[] Justification: The MyoSuite is cited as the simulator embedded within MuJoCo with a URL. The participants' solutions are cited and referenced accordingly. The license is available on the MyoSuite GitHub repository. Only publicly available data are used. 

\item {\bf New assets}
    \item[] Question: Are new assets introduced in the paper well documented and is the documentation provided alongside the assets?
    \item[] Answer: \answerYes{} 
    \item[] Justification: All assets are available from the \href{https://github.com/MyoHub/myosuite}{MyoSuite GitHub repository} including license, model and code.

\item {\bf Crowdsourcing and research with human subjects}
    \item[] Question: For crowdsourcing experiments and research with human subjects, does the paper include the full text of instructions given to participants and screenshots, if applicable, as well as details about compensation (if any)? 
    \item[] Answer: \answerNA{}
    \item[] Justification: No human subjects were involved.
    

\item {\bf Institutional review board (IRB) approvals or equivalent for research with human subjects}
    \item[] Question: Does the paper describe potential risks incurred by study participants, whether such risks were disclosed to the subjects, and whether Institutional Review Board (IRB) approvals (or an equivalent approval/review based on the requirements of your country or institution) were obtained?
    \item[] Answer: \answerNA{}
    \item[] Justification: No human subjects were involved.

\item {\bf Declaration of LLM usage}
    \item[] Question: Does the paper describe the usage of LLMs if it is an important, original, or non-standard component of the core methods in this research? Note that if the LLM is used only for writing, editing, or formatting purposes and does \emph{not} impact the core methodology, scientific rigor, or originality of the research, declaration is not required.
    \item[] Answer: \answerNA{}
    \item[] Justification: The core method development in this research does not involve LLMs as any important, original, or non-standard components.

\end{enumerate}

\newpage
\appendix

\section{Competition Details}
The competition ran from July 21st to November 8th on EvalAI \url{https://eval.ai/web/challenges/challenge-page/2628/overview}, with a final workshop in the NeurIPS 2025 conference competition: MyoSymposium (\url{https://sites.google.com/view/myosuite/myochallenge/myochallenge-2025}). The workshop allowed winners from both tracks to present their solutions and bring together researchers and scholars in the field of biomechanics, ML, neuroscience, and sports science.

All environments and task configurations can be cloned from the MyoChallenge25 GitHub template (\url{https://github.com/MyoHub/myochallenge_2025eval}). The challenge was hosted on EvalAI (\url{https://eval.ai}), where the policies uploaded by participants underwent automated evaluation for a public scoreboard. Final scores reflect the average performance across 100 unseen trials to ensure robustness against multiple seeds and unseen environmental variations. 

\section{EvalAI Computation Resources}\label{appendix: comput}
The MyoChallenge utilizes EvalAI (\cite{EvalAI2019}) for its evaluation platform and deployment support. Its computational infrastructure is powered by Amazon Web Services (AWS), as detailed in the following list:
\begin{itemize}
    \item AWS EC2 - c5.4xlarge (16 CPU, 32 GB RAM)
    \item AWS EBS - gp2 (17 GB)
    \item AWS ECS for Kubernetes
    \item AWS ECR
\end{itemize}

\section{Environment Details} \label{appendix:env}
In the evaluation on EvalAI, a hidden set of parameters is used to ensure fairness across submissions. All efforts from the EvalAI platform are recomputed using the following formula:

$$\text{Effort} = \frac{1}{n_a} \sum_{i=1}^{n_a} a_i^2$$

\subsection{Table Tennis Track} \label{app:tt}

\paragraph{Table Tennis Baseline} The table tennis baseline uses a 12 CPU with one 5090 RTX GeForce. The baseline model implements PPO using the $\text{stable\_baselines3}$ library. The policy network follows an ActorCriticPolicy architecture with an MLP comprising three hidden layers of sizes [1024, 512, 512], respectively, employing the SiLU activation function. The agent was trained for a total of 10 million timesteps, collecting 2,048 steps per rollout across six parallel environments, followed by ten update epochs with a batch size of 64. Key algorithmic parameters include a discount factor ($\gamma$) of 0.99, a GAE lambda of 0.95, an entropy coefficient of 0.001 to encourage exploration, a value function coefficient of 0.5, and a maximum gradient norm of 0.5 for gradient clipping. Both the learning rate and the PPO clipping range follow a linear decay schedule, annealing from initial values of 1.0 and 0.2, respectively, to zero over the course of training.

\paragraph{Termination Condition} An episode terminates when any of the following conditions are met: (1) the maximum episode duration is exceeded; (2) the ball's vertical position drops below a critical height threshold ($z < 0.3$,m); (3) the task is successfully solved; or (4) the ball exhibits an invalid contact trajectory. An invalid trajectory is strictly defined as a Double Touch (contacting the paddle multiple times), an Invalid Bounce (bouncing twice or rolling on the agent's table side), a Missed Strike (landing on the opponent's side without paddle contact), or an Incomplete Play (failing to land on the opponent's side after a valid strike).

\begin{table}[htb]
  \centering
  \caption{Unvaried Object properties for the Table Tennis.}
  \label{table:task_object_properties}
  \renewcommand{\arraystretch}{1.2}
  \begin{tabular}{@{} l l @{}}
    \toprule
    \textbf{Object} & \textbf{Properties} \\
    \midrule
    \multicolumn{2}{@{}l}{\textbf{Table Tennis Task}} \\
    \quad Table Top (Total) & $2.74 \times 1.52 \times 1.59$\,m$^3$ \\
    \quad Table Top (Each Side) & $1.37 \times 1.52 \times 1.59$\,m$^3$ \\
    \quad Net & $0.01 \times 1.825 \times 0.305$\,m$^3$ \\
    \quad Paddle Handle & Radius: $1.6$\,cm, Height: $5.1$\,cm \\
    \quad Paddle Face & Radius: $9.3$\,cm, Height: $2$\,cm \\
    \quad Ball Radius & $2$\,cm \\
    \quad Ball Mass & $2.7$\,g \\
    \quad Ball Inertia & $7.2 \times 10^{-7}$\,kg$\cdot$m$^2$ \\
    \bottomrule
  \end{tabular}
\end{table}

\begin{table}[htb]
  \centering
  \caption{Changing Object properties across phases for table tennis tasks. During the first phase, all initial starting criteria are kept constant except for a slight change in ball starting position. During the second phase, we feed in variations in terms of mass, friction for the ball and the paddle. In addition, now the incoming trajectory could span towards the entire side of the model's table at various speeds. It is not guaranteed that the model can always return the table tennis successfully.}
  \label{table:tt_phase_comparison}
  \begin{tabular}{@{} l c c c @{}}
    \toprule
    \textbf{Property} & \textbf{Phase 1} & \textbf{Phase 2} & \textbf{Evaluation} \\
    \midrule
    \textbf{Paddle Mass} & 150\,g & 100--150\,g & 90--200\,g \\
    \midrule
    \textbf{Ball Friction} & & & \\
    \hspace{1em} Sliding & 1.0 & $[0.9, 1.1]$ & $[0.9, 1.1]$ \\
    \hspace{1em} Torsional & 0.005 & $[0.004, 0.006]$ & $[0.004, 0.006]$ \\
    \hspace{1em} Rolling & 0.0001 & $[0.00001, 0.00003]$ & $[0.00001, 0.00003]$ \\
    \midrule
    \textbf{Ball Start Pos.} & & & \\
    \hspace{1em} $x$ & $[-1.25, -1.20]$ & $[-1.25, -0.50]$ & $[-1.25, -0.40]$ \\
    \hspace{1em} $y$ & $[-0.50, -0.45]$ & $[-0.50, 0.50]$ & $[-0.50, 0.50]$ \\
    \hspace{1em} $z$ & $[1.40, 1.50]$ & $[1.40, 1.50]$ & $[1.30, 1.55]$ \\
    \midrule
    \textbf{Ball Velocity} & & & \\
    \hspace{1em} $v_x$ & 5.6 & \multicolumn{2}{c}{} \\
    \hspace{1em} $v_y$ & 1.6 & \multicolumn{2}{c}{Varies\textsuperscript{*}} \\
    \hspace{1em} $v_z$ & 0.1 & \multicolumn{2}{c}{} \\
    \bottomrule
  \end{tabular}
  
  {\footnotesize \textsuperscript{*}Guaranteed to land on the agent's side.\par}
\end{table}

\begin{table}[htb]
  \centering
  \caption{Observation Space for Table Tennis Task}
  \renewcommand{\arraystretch}{1.2}
  \begin{tabular}{@{} l l c @{}}
    \toprule
    \textbf{Description} & \textbf{Component} & \textbf{Count} \\
    \midrule
    Pelvis Position & \texttt{pelvis\_pos} & 3 \\
    Joint Positions & \texttt{body\_qpos} & 58 \\
    Joint Velocities & \texttt{body\_vel} & 58 \\
    Ball Position & \texttt{ball\_pos} & 3 \\
    Ball Velocity & \texttt{ball\_vel} & 3 \\
    Paddle Position\textsuperscript{*} & \texttt{paddle\_pos} & 3 \\
    Paddle Velocity & \texttt{paddle\_vel} & 3 \\
    Paddle Orientation & \texttt{paddle\_ori} & 3 \\
    Paddle Reaching Error (see below) & \texttt{reach\_err} & 3 \\
    Muscle Activations & \texttt{muscle\_activations} & 273 \\
    Touching Information \textsuperscript{**} & \texttt{touching\_info} & 6 \\
    \bottomrule
  \end{tabular}
  \label{table:tennis_observation_space}

  {\footnotesize 
  \textsuperscript{*}This measures the spatial distance between the paddle and the table tennis object.\\
  \textsuperscript{**}This consists of six boolean indicators representing the ping-pong ball’s contact status with various objects in the environment: \textbf{Paddle} (contact with the paddle), \textbf{Own} (the agent's table side), \textbf{Opponent} (the opponent table side), \textbf{Ground}, \textbf{Net}, and \textbf{Env} (any other part of the environment).}
\end{table}

\subsection{Soccer Track} \label{app: soccer}

\paragraph{Goal Keeper Behavior} A simple goal keeper, created with Mujoco native shapes, acts as a goal keeper to increase the challenge of the task for trained policies. There are 3 different goal keeper behaviors, where the environment randomly selects during evaluation. They are:
\begin{enumerate}
    \item \textbf{Random static location along the goal line:} Goal keeper appears at a random location along the goal line at the start of each episode, and remains stationary for the entire episode
    \item \textbf{Random movement along the goal line:} Goal keeper moves randomly along the goal line throughout the episode, at a randomized velocity
    \item \textbf{Ball tracking behavior along the goal line:} Goal keeper tracks the position of the ball, but does not leave the goal line 
\end{enumerate}
Goal keeper behaviors are parameterized by the velocity they are moving along the goal line. Goal keeper velocity is randomized between 1m/s to 5m/s at the start of each episode.

\begin{table}[htb]
  \centering
  \caption{Unvaried Object properties for the Soccer tasks.}
  \label{table:task_object_properties_soccer}
  \renewcommand{\arraystretch}{1.2}
  \begin{tabular}{@{} l l @{}}
    \toprule
    \textbf{Object} & \textbf{Properties} \\
    \midrule
    \multicolumn{2}{@{}l}{\textbf{Soccer Task}} \\
    \quad Soccer Net & Width: 7.32\,m, Height: 2.50\,m \\
    \quad Ball Radius & 0.117\,m \\
    \quad Ball Mass & 450\,g \\
    \bottomrule
  \end{tabular}
\end{table}

\begin{table}[htb]
  \centering
  \caption{Comparison of starting criteria across challenge phases for the soccer task.}
  \label{table:soccer_phase_comparison}
  \begin{tabular}{@{} l c c c @{}}
    \toprule
    \textbf{Property} & \textbf{Phase 1} & \textbf{Phase 2} & \textbf{Evaluation} \\
    \midrule
    \textbf{Ball Start Pos.} &\multicolumn{3}{c}{Fixed} \\
    \midrule
    \textbf{Agent Start Pos.} & & & \\
    \hspace{1em} $x$ & 39.0 & $[38.0, 39.0]$ & $[36.5, 39.0]$ \\
    \hspace{1em} $y$ & 0.0 & $[\pm 1.0]$ & $[\pm 2.5]$ \\
    \midrule
    \textbf{Agent Orientation} & \multicolumn{3}{c}{Facing goal, in front of the ball} \\
    \midrule
    \textbf{Position Noise}\textsuperscript{*} & 0.0 & 1.0 & 2.5 \\
    \midrule
    \textbf{Joint Noise}\textsuperscript{**} & 0.0 & 0.02 & 0.03 \\
    \midrule
    \textbf{Goalkeeper}\textsuperscript{$\dagger$} & 100\% S & \begin{tabular}[c]{@{}c@{}} 60\% S, 30\% R, 10\% T \\ $v \in [1, 5]$ \end{tabular} & \begin{tabular}[c]{@{}c@{}} 10\% S, 45\% R, 45\% T \\ $v \in [3.5, 5.5]$ \end{tabular} \\
    \midrule
    \textbf{Max Time} & 20\,s & 10\,s & 10\,s \\
    \bottomrule
  \end{tabular}
  \vspace{0.2cm}
  
  {\footnotesize 
  \textsuperscript{*}Shifts base $x$ by $[-v, 0]$ and $y$ by $[-v, v]$.\\
  \textsuperscript{**}Adds uniform noise $\in [-\mu, \mu]$ to initial joint angles.\par
  \textsuperscript{$\dagger$} S: Stationary, R: Random, T: Tracking soccer ball.}
\end{table}

\begin{table}[htb]
  \centering
  \caption{Observation space components for the soccer task.}
  \label{tab:obs_space_soccer}
  \begin{tabular}{@{} l l c @{}}
    \toprule
    \textbf{Description} & \textbf{Component} & \textbf{Count} \\
    \midrule
    Ball Position & \texttt{ball\_pos} & 3 \\
    4 Position Coords (bounding goal area) & \texttt{goal\_bounds} & 12 \\
    Muscle Activations & \texttt{act} & 290 \\
    Joint Angles & \texttt{internal\_qpos} & 46 \\
    Joint Velocities & \texttt{internal\_qvel} & 46 \\
    Foot Position (Right) & \texttt{r\_toe\_pos} & 3 \\
    Foot Position (Left) & \texttt{l\_toe\_pos} & 3 \\
    Body COM in world frame\textsuperscript{*} & \texttt{model\_root\_pos} & 7 \\
    Body COM vel in world frame\textsuperscript{*} & \texttt{model\_root\_vel} & 6 \\
    Goal Keeper Position  & \texttt{goalkeeper\_pos} & 2 \\
    \bottomrule
  \end{tabular}
  
  \vspace{0.2cm}
  {\footnotesize \textsuperscript{*}The body COM position is represented with a free joint, hence the 7 dimensions: $[x, y, z, q_x, q_y, q_z, q_w]$. Similarly, the body COM velocity is represented with 6 dimensions: $[v_x, v_y, v_z, \alpha, \beta, \gamma]$.\par}
\end{table}

\clearpage
\section{Table Tennis First Place  - Acting AI Full Solution} \label{app:actingai}

\subsection{Reward Design}
Our total reward function $R_{total}$ integrates terms for dense grasping reward, dense tracking reward and sparse task finishing reward:
\begin{equation}
R_{total} = w_g R_{grasp} + w_c R_{tracking} + w_s R_{task}
\end{equation}

The dense grasping reward encourages the agent to maintain a stable grip on the paddle throughout the episode. It consists of three terms: the relative position error between the paddle and the grasp site, the relative orientation error, and a penalty for finger opening:

\begin{itemize}
    \item \textbf{relative position:} $r_{rel\_p} = \exp\!\left(-8 \left\|\mathbf{p}_{\text{rel}} - \mathbf{p}_{\text{rel}}^{init}\right\|_2\right)$
    \item \textbf{relative rotation:} $r_{rel\_r} = \exp\!\left(-4 \arccos\!\left(\left|\mathbf{q}_{\text{rel}} \cdot \mathbf{q}_{\text{rel}}^{init}\right|\right)\right)$
    \item \textbf{finger open:} $r_{fin\_open} = \exp\!\left(-5 \sum_{i=0}^{4} \left\|\mathbf{p}_{\text{fin}_i} - \mathbf{p}_{\text{palm}}\right\|_2\right)$
\end{itemize}
where $\mathbf{p}_{\text{rel}} = \mathbf{R}_{\text{grasp}}^{\top}(\mathbf{p}_{\text{paddle}} - \mathbf{p}_{\text{grasp}})$ is the paddle position expressed in the grasp frame.
$\mathbf{q}_{\text{rel}} = \mathbf{q}_{\text{grasp}}^{-1} \otimes \mathbf{q}_{\text{paddle}}$ is the relative orientation quaternion between the paddle and the grasp site. $\mathbf{p}_{\text{rel}}^{init}$ and $\mathbf{q}_{\text{rel}}^{init}$ are their values at the beginning of each episode. And $\mathbf{p}_{\text{fin}_i}$, $\mathbf{p}_{\text{palm}}$ denote the positions of the $i$-th fingertip and the palm center, respectively.

The dense tracking reward guides the agent to move the paddle towards the planned
hit position with the correct orientation before contact, and to match the target
paddle velocity at the moment of contact:

\begin{itemize}
    \item \textbf{reach distance:}
    $r_{\text{reach}} = \exp\!\left(-4 \left\|\mathbf{p}_{\text{paddle}} - \mathbf{p}_{\text{hit}}\right\|_2\right), \quad t < t_{\text{hit}}$

    \item \textbf{paddle orientation:}
    $r_{\text{ori}} = \exp\!\left(-2 \left\|\mathbf{q}_{\text{paddle}} - \mathbf{q}_{\text{hit}}\right\|_2\right), \quad t < t_{\text{hit}}$

    \item \textbf{paddle velocity:}
    $r_{\text{vel}} = \exp\!\left(-\left\|\mathbf{v}_{\text{paddle}} - \mathbf{v}_{\text{hit}}\right\|_2\right), \quad t = t_{\text{hit}}$
\end{itemize}

where $\mathbf{p}_{\text{hit}}$, $\mathbf{q}_{\text{hit}}$ and $\mathbf{v}_{\text{hit}}$ are the planned hitting position,
orientation and velocity computed by the physics-based planner.
$t_{\text{hit}}$ is the planned hit time. The reach and orientation terms are
active only before $t_{\text{hit}}$ and before any ball--paddle contact occurs,
while the velocity term is a sparse signal triggered only at $t = t_{\text{hit}}$.

The sparse task finishing reward provides goal-oriented feedback upon successful
task completion. It consists of three terms:

\begin{itemize}
    \item \textbf{hit success:}
    $r_{\text{hit}} = \mathbf{1} \text{ if paddle successfully contacts the ball else } \mathbf{0}$

    \item \textbf{land success:}
    $r_{\text{land}} = \mathbf{1} \text{ if the predicted land position is within the valid range else } \mathbf{0}$

    \item \textbf{landing accuracy:}
    $r_{\text{acc}} = \exp\!\left(-0.5\left\|\hat{\mathbf{p}}_{\text{land}} - \mathbf{p}_{\text{opp\_center}}\right\|_2\right)$
\end{itemize}

where $\hat{\mathbf{p}}_{\text{land}}$ is the analytically predicted landing position of the ball computed from its current position and velocity after the hit, and $\mathbf{p}_{\text{opp\_center}}$ is the center of the opponent half table. The landing accuracy term $r_{\text{acc}}$ is only activated when the ball is
predicted to land within opponent range.

\subsection{Computer Resources:}
All experiments were performed on a local workstation featuring an AMD Ryzen 9 7950X
16-core processor and a single NVIDIA GeForce RTX 4090 GPU with 24\,GB of VRAM.
A policy trained from scratch over 300 million timesteps cost 51 hours of training time.

\subsection{Training Details:} The policy was trained using the Proximal Policy Optimization (PPO) algorithm. The hyperparameters are shown in Table~\ref{tab:acting_ai_hyperparams}.

\begin{table}[h]
\centering
\caption{Hyperparameters for PPO Training.}
\label{tab:acting_ai_hyperparams}
\begin{tabular}{lll}
\toprule
\textbf{Category} & \textbf{Hyperparameter} & \textbf{Value} \\
\midrule
\multirow{4}{*}{PPO} 
  & Learning rate         & $3\times10^{-4}$ (linear decay) \\
  & Steps per update      & 200 \\
  & Batch size            & 12800 \\
  & Total timesteps       & $10^{8}$ \\
\midrule
\multirow{3}{*}{Network}
  & Architecture          & MLP [1024, 1024, 512, 512, 256, 256] \\
  & Activation function   & SiLU \\
\midrule
\multirow{2}{*}{Environment}
  & Parallel environments & 64 \\
  & Observation normalization & enabled \\
\midrule
\multirow{7}{*}{Reward weights}
  & $w_{\text{rel\_p}}$         & 1 \\
  & $w_{\text{rel\_r}}$         & 1 \\
  & $w_{\text{fin\_open}}$        & 1 \\
  & $w_{\text{reach}}$            & 5 \\
  & $w_{\text{paddle\_ori}}$      & 5 \\
  & $w_{\text{vel}}$              & 50 \\
  & $w_{\text{hit\_success}}$       & 100 \\
  & $w_{\text{land\_success}}$       & 100 \\
  & $w_{\text{land\_acc}}$       & 100 \\
\bottomrule
\end{tabular}
\end{table}

\subsection{Code Availablity}
The code of Diff-Muscle is available at \url{https://github.com/TaoshuaiZ/Diff-Muscle}.

The submission repository is publicly available at \url{https://github.com/TaoshuaiZ/Team_ActingAI_Submit}.

\clearpage
\section{Table Tennis Second Place  - BioSyn Full Solution} \label{app:biosyn}

\subsection{Multivariate Gaussian Noise Definition}
To improve the exploration efficiency within the redundant muscle action space, the exploration noise in BioSyn is defined as a multivariate normal distribution associated with specific muscle synergy weights:
\begin{equation}
\begin{split}
    \pi(\boldsymbol{a}|\boldsymbol{s}) &\sim \mathcal{N} \bigl( \mathrm{W}_a x_a \oplus \mathrm{W}_h x_h \oplus \mathrm{W}_t x_t \oplus \mathrm{W}_p x_p,\; \Sigma_{syn} + \Sigma_o \bigr) \\
    \Sigma_{syn} &= \mathrm{diag}\bigl( \mathrm{W}_a \Sigma_a {\mathrm{W}_a}^{\top},\; \mathrm{W}_h \Sigma_h {\mathrm{W}_h}^{\top},\; \mathrm{W}_t \Sigma_t {\mathrm{W}_t}^{\top},\; \mathrm{W}_p \Sigma_p {\mathrm{W}_p}^{\top} \bigr)
\end{split}
\end{equation}
where $\mathrm{W}_h$, $\mathrm{W}_a$, $\mathrm{W}_t$, and $\mathrm{W}_p$ denote the synergy weights for the hand, arm, torso, and pelvis, respectively. $x_h$, $x_a$, $x_t$, and $x_p$ denote the corresponding synergy modulations. $\Sigma_o$ denotes the original exploration noise used in reinforcement learning. $\Sigma_h$, $\Sigma_a$, $\Sigma_t$, and $\Sigma_p$ denote the synergy-specific exploration noise for the hand, arm, torso, and pelvis, respectively.

\subsection{Reward Design}
The reward function was designed to guide the BioSyn agent to execute the striking commands issued by the high-level planner. The key terms in the reward design are described as follows:
\begin{itemize} 
    \item Hand–handle distance: $r_{h\_h} = \exp\!\left(-5 \left\|\mathbf{p}_{\text{hand}} - \mathbf{p}_{\text{handle}}\right\|_2\right)$
    \item Paddle position: $r_{paddle} = \exp\!\left(-5 \left\|\mathbf{p}_{\text{paddle}} - \mathbf{p}_{\text{target}}\right\|_2\right)$
    \item Paddle orientation: $r_{\mathrm{rot}} = \exp\Big(-10 \, \big\| \mathrm{vec}\big(\mathbf{q}_{\text{target}} \otimes \mathbf{q}_{\text{paddle}}^{-1}\big) \big\|_2 \Big) $
    \item Paddle drop: $r_{\text{drop}} = -\mathbf{1} \text{ if the paddle is dropped from the hand else } \mathbf{0}$
    \item Hit success: $r_{\text{hit}} = \mathbf{1} \text{ if paddle successfully contacts the ball else } \mathbf{0}$
    \item paddle velocity: $r_{\text{vel}} = \exp\!\left(-\left\|\mathbf{v}_{\text{paddle}} - \mathbf{v}_{\text{hit}}\right\|_2\right), \quad t_{\text{hit}}-0.03s < t < t_{\text{hit}}+0.03s$
\end{itemize}  

where $\mathbf{p}_{\text{hand}}, \mathbf{p}_{\text{handle}}, \mathbf{p}_{\text{paddle}}, \mathbf{p}_{\text{target}} $
are the positions of the hand, the paddle handle, the paddle, and the desired striking target, respectively. $\mathbf{q}_{\text{paddle}}, \mathbf{q}_{\text{target}}$ are the orientations of the current paddle and the desired striking target, represented as unit quaternions. $\mathrm{vec}(\mathbf{q})$ denotes the vector part (last three components) of quaternion $\mathbf{q}$. 
$\otimes$ represents quaternion multiplication.  
$\mathbf{v}_{\text{paddle}}, \mathbf{v}_{\text{hit}}$ are the current paddle velocity and the desired hitting velocity.  
$t_{\text{hit}}$ denotes the desired hitting time. 
The indicator functions in $r_{\text{drop}}$ and $r_{\text{hit}}$ take values of 1 or 0 depending on whether the paddle is dropped or successfully contacts the ball, respectively. The detailed curriculum reward weights are provided in Sec.~\ref{biosyn:code}.

\subsection{Weakness-Aware Curriculum}
Inspired by table tennis training principles, team BioSyn proposes a Weakness-Aware Table Tennis Curriculum. During the early stage of training, the curriculum primarily emphasizes learning in the default hitting environment. In the later stages, the hitting performance of the policy is systematically evaluated to identify weak regions in the striking space. Based on this evaluation, the environment sampling strategy at subsequent curriculum stage is adapted to increase the probability of sampling balls from these weak regions, enabling the policy to progressively improve its deficient hitting skills.

\subsection{Computer Resources}
Team BioSyn trained their solution on a single Ubuntu server featuring two AMD EPYC 9654 CPUs (96 cores each), an NVIDIA GeForce RTX 4090 GPU with 24GB of memory, and 256GB of DDR5 RAM. Training required nearly 15 days to complete 1 billion steps.

\subsection{Training Details}
The key hyperparameters for reinforcement learning in BioSyn are summarized in Table~\ref{tab:hyperparams_biosyin}.
\begin{table}[h]
\centering
\caption{Hyperparameters for the BioSyn training framework.}
\label{tab:hyperparams_biosyin}
\begin{tabular}{@{}ll@{}}
\toprule
\textbf{Hyperparameter} & \textbf{Value} \\ \midrule
Number of Parallel Environments & 180 \\
Network Archiecture & LSTM[256]+MLP[256, 256, 150] \\
Number of Hand Synergies & $20$ \\
Number of Arm Synergies & $20$ \\
Number of Torso Synergies & $100$ \\
Actication Function & Relu \\
Learning Rate & $1\times10^{-5}$ \\
Updates per Iteration & 5 \\
Entropy Coefficient & $1\times10^{-5}$ \\
Batch Size & $2048$ \\
Steps per Update & $512$ \\ 
\bottomrule
\end{tabular}
\end{table}

\subsection{Code availability}
\label{biosyn:code}

The BioSyn detail training code is available at \url{https://github.com/Siyuan-Liu99/BioSynTT}.

The submission repository is publicly available at \url{https://github.com/Siyuan-Liu99/myochallenge2025eval_BioSyn}.

\clearpage
\section{Table Tennis Third Place  - LNSGroup Full Solution}\label{app:lnsgroup}

\subsection{Reward Design}
Our reward ($R_{total}$) decomposes the task into dense geometric tracking, biomechanical kinematics, sparse task completion, and rule penalties:
\begin{align*}
    R_{total} = w_{track}R_{track} &+ w_{quat}R_{quat} + w_{torso}R_{torso} + \\ w_{palm}R_{palm} &+ w_{hit}R_{hit} + w_{success}R_{success} + w_{own}R_{own}
\end{align*}

\noindent 1. Dense Geometric Tracking: Guides the paddle towards the predicted physical targets.
\begin{align*}
    \text{Tracking} &: R_{track} = \exp\left(-5 \|\mathbf{p}_{paddle} - \mathbf{p}_{hit}\|_2\right) \\
    \text{Orientation} &: R_{quat} = \frac{1}{2}\left (  \frac{\mathbf{d}_{paddle} \cdot \mathbf{d}_{hit}}{\|\mathbf{d}_{paddle}\| \|\mathbf{d}_{hit}\|}  + 1 \right )
\end{align*}
where $\mathbf{p}_{paddle}$ and $\mathbf{d}_{paddle}$ denote the position and normal direction vector of the paddle surface, while $\mathbf{p}_{hit}$ and $\mathbf{d}_{hit}$ represent the optimal striking position and desired striking direction predicted by the physical prior model, respectively.

\noindent 2. Biomechanical Kinematics: Introduces an upright torso reward ($R_{torso}$) and a palm-to-handle proximity reward ($R_{palm}$) to prevent spine collapse and paddle decoupling independently, ensuring a stable and continuous mechanical base.

\noindent 3. Sparse Task Completion: Functions as the core task incentive.
\begin{itemize}
    \item Hit Sparse: $1.0$ when the paddle successfully strikes the ball.
    \item Success: $1.5$ if the ball lands inside an optimal rectangular target zone on the opponent's side, $1.0$ for any valid opponent table hit, and $0.0$ otherwise.
\end{itemize}

\noindent 4. Rule-based Penalty: 
Penalizes the agent if the ball bounces on its own side after the expected hitting time $t_{hit}$ ($R_{own} = -1.0$).

\subsection{Computational Resources}
Training experiments were conducted on a workstation equipped with a single NVIDIA 80GB A100 GPU and two Intel(R) Xeon(R) Gold 6348 CPUs. Each RL training run was initialized from scratch for 100 million steps, taking approximately 12 hours.

\subsection{Training and Evaluation Details}
The key hyperparameters for the reinforcement learning framework are summarized in Table~\ref{tab:hyperparams_lns}.
\begin{table}[h]
\centering
\caption{Hyperparameters for the CrossQ training framework.}
\label{tab:hyperparams_lns}
\begin{tabular}{@{}ll@{}}
\toprule
\textbf{Hyperparameter} & \textbf{Value} \\ \midrule
Number of Vectorized Environments & 64 \\
Hidden Dimension & 1024 \\
Hidden Layers & 2 \\
Base Learning Rate & $5 \times 10^{-4}$ (Linear decay to 20\%) \\
Entropy Learning Rate ($\alpha$) & $7 \times 10^{-3}$ \\
Buffer Size & $1\,000\,000$ \\
Updates per Iteration & 1 \\
\bottomrule
\end{tabular}
\end{table}

\subsection{Code availability}
\label{lnsgroup:code}

The training and submission code is available at \url{https://github.com/LNSGroup/MyoChallenge2025-LNSGroup}.

\newpage
\section{Soccer First Place  - Servette MyoClub Full Solution} \label{app:servette}

\subsection{From Motion Imitation to Soccer Kicking (Teacher Policy)}
\label{sec:teacher}

KINESIS' motion imitation policy tracks reference motions via target keypoint
observations. To obtain a target-reaching policy, a simple
observation modification was applied: the root (pelvis) position in the target
observation was set to the desired target location, while all other target
keypoints were set to the current positions of their respective body parts, following
prior work~\cite{simos2025kinesis,chiappa2025arnold0}. Motion tracking reward components were also replaced with target-reaching rewards inspired by Won et
al.~\cite{won2022physics}. This reward formulation directs the policy to walk
toward the specified target while maintaining a natural gait for the rest of the
body.

The target position was uniformly sampled across a
$2 \times 2\,\text{m}^2$ area around the agent, until the policy successfully
reached the entire sampled region. Subsequently, the agent was transferred to the penalty-kick environment,
where the target position was set to the ball's location, and the agent was initialized at a default upright position. The target-reaching policy was fine-tuned with a reward
based on ball velocity along the goal direction. The kicking policy was trained
and evaluated in an environment without joint-level noise and with only minor
position randomization. This produced a teacher policy that could reliably walk
towards the ball and kick it, achieving a goal conversion rate of approximately
25\% in the MyoChallenge evaluation setting. For comparison, training a kicking
policy from scratch with the same reward fails entirely, as the agent cannot
learn to balance, walk, and kick simultaneously in the 290-dimensional muscle
action space.

\subsection{Retargeting to the MyoChallenge environment}
\label{sec:adaptation}

The soccer kicking policy from the previous stage operates on KINESIS
observations (453 dimensions), whereas the MyoChallenge environment provides a
different observation space comprising: \texttt{internal\_qpos},
\texttt{internal\_qvel}, \texttt{grf}, \texttt{torso\_angle},
\texttt{ball\_pos}, \texttt{model\_root\_pos}, \texttt{model\_root\_vel},
\texttt{muscle\_length}, \texttt{muscle\_velocity}, and
\texttt{muscle\_force}. To bridge this gap, a four-step adaptation
pipeline was employed.

\paragraph{Offline trajectory collection with success filtering.}

The KINESIS soccer kicking policy (teacher) was deployed in the MyoChallenge
environment with domain randomization (random joint-level noise and random
initial positions) and trajectory rollouts were collected. Trajectories were filtered based on success rate, retaining only episodes with successful kicking behavior.
At each timestep, both the KINESIS-format and MyoChallenge-format
observations we recorded, along with the teacher's muscle controls.

\paragraph{Student policy pre-training.}

A student policy, which takes MyoChallenge observations as input, was pre-trained on the filtered offline trajectories via supervised learning (behavioral cloning
on the collected dataset). This provided a stable warm-start initialization for
the subsequent on-policy stage.

\begin{figure}
    \centering
    \includegraphics[width=0.6\linewidth]{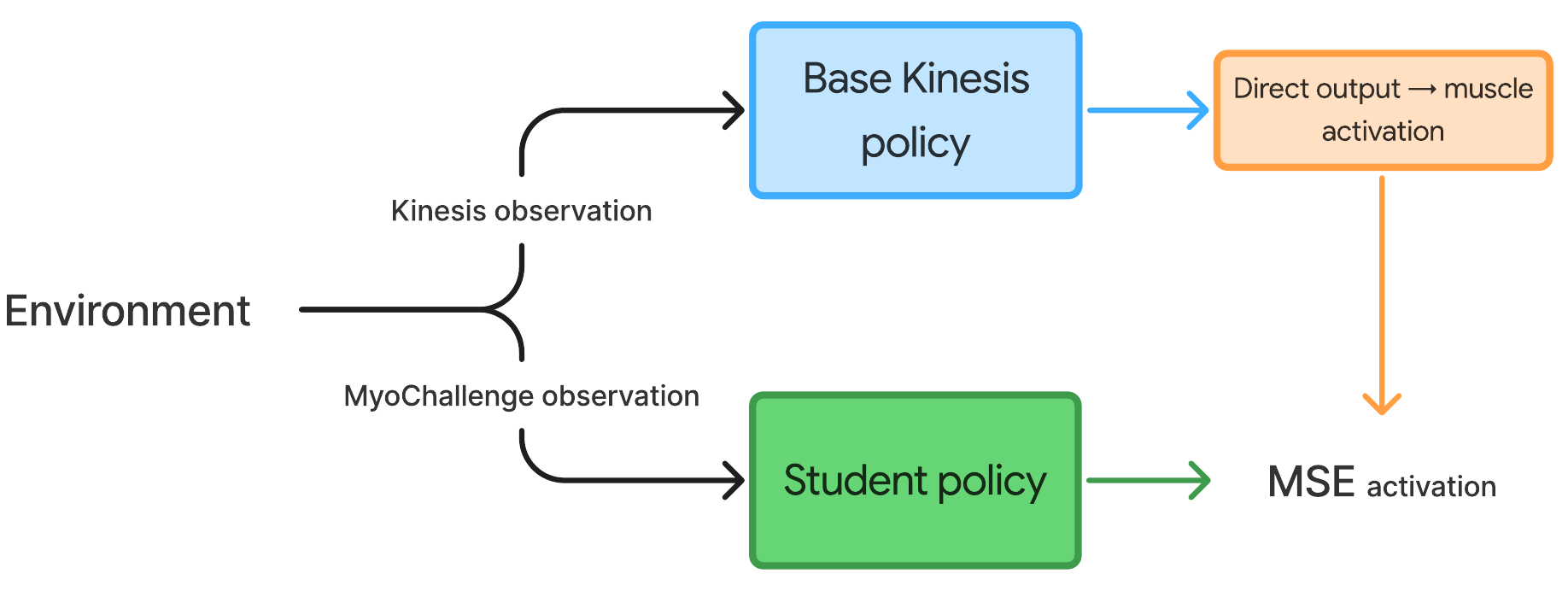}
    \caption{Schematic diagram of online behavior cloning.}
    \label{fig:obc}
\end{figure}

\paragraph{On-policy behavior cloning (Online DAgger, OBC~\cite{chiappa2025arnold0}).}
The pre-trained student was further refined through on-policy behavior cloning~ (Figure~\ref{fig:obc}).
The environment was modified to simultaneously generate both KINESIS and
MyoChallenge observations from the same simulator state. The KINESIS observation
was routed to the teacher policy to produce target muscle controls; the
MyoChallenge observation was provided to the student policy. The student was
trained to minimize the mean squared error (MSE) between its predicted
activations and the teacher's activations. This was performed on-policy: the
student generated its own rollouts and the teacher provided target activations at
each visited state, avoiding the covariate shift inherent in offline-only
behavioral cloning, similar to DAgger~\cite{ross2011reduction}. Domain
randomization (random joint-level noise and random initial positions) was applied
throughout this stage. After OBC, the student policy achieved a goal conversion rate of
34\%, exceeding the teacher policy's 25\%, with a MSE of approximately
0.019.

\paragraph{Reinforcement learning fine-tuning.}
The behavior-cloned student policy was fine-tuned with PPO using conservative
updates to preserve the learned locomotion prior. Specifically, three
techniques were considered: (1) reduced PPO clipping parameter $\varepsilon$ to constrain policy
updates; (2) reduced actor learning rate; and (3) an additional behavior cloning
regularization term, yielding the combined loss:

\begin{equation}
  \mathcal{L} = \mathcal{L}_{\text{PPO}} + \lambda_{\text{BC}}\,\mathcal{L}_{\text{BC}},
  \label{eq:loss}
\end{equation}

\noindent where $\lambda_{\text{BC}}$ controls the strength of the behavior
cloning regularizer. Early termination conditions were applied when the agent lost balance or when the ball was clearly out of scoring range.

\subsection{Reward Design}
\label{sec:rewards}

\paragraph{KINESIS motion imitation reward.}
The base reward combined four weighted terms: position tracking (weight 0.6,
$k=200$), velocity tracking (weight 0.2, $k=5$), energy regularization (weight
0.1, $\ell_1+\ell_2$ on muscle activations), and upright reward (weight 0.1,
$k=3$).

\paragraph{KINESIS target reaching reward.}
Position and velocity tracking components were replaced with
\begin{equation}
  r_{\text{track}} = \delta_{t-1} - \delta_t, \qquad
  r_{\text{success}} = \mathbf{1}\!\left[\delta_t < 0.1\right],
  \label{eq:reach}
\end{equation}
where $\delta$ denotes the distance between the root and the target position.

\paragraph{Soccer kicking reward.}
The kicking policy was fine-tuned with a reward based on the forward ball
velocity $v_x$ (along the goal direction) and the lateral ball velocity $v_y$:
\begin{equation}
  r_{\text{kick}} =
  \begin{cases}
    v_x &
      \text{if } v_x \geq \epsilon_v
      \;\text{ and }\;
      \dfrac{v_x}{|v_y| + \epsilon_r} \geq 1, \\[6pt]
    0 & \text{otherwise,}
  \end{cases}
  \label{eq:kick}
\end{equation}
where $\epsilon_v = 10^{-2}\,\text{m/s}$ is a minimum forward-speed threshold, and
$\epsilon_r = 10^{-6}$ is a small regularizer that prevents division by zero. The two gate conditions enforce that (i) the ball is moving forward with non-negligible speed, and (ii) the
forward-to-lateral velocity ratio satisfies $v_x / |v_y| \geq 1$, which restricts rewarded kicks to a cone of $\pm 45^{\circ}$ around the goal direction. When both conditions are met, the reward equals the forward ball
velocity, encouraging maximum kick power.
Angular kicks toward the goal corners emerged as a natural consequence of goalkeeper movement randomness, the randomly off-center initialization of the agent, and the goal-scoring reward, without explicit reward shaping for kick angle.

\subsection{Results}
\label{sec:servette-results}

The pre-trained KINESIS policy managed to move towards the ball and initiate contact with it in a zero-shot manner, thus already solving the hard problem of balance and locomotion (Figure~\ref{fig:viz}). After implementing the full training pipeline, the policy consistently converted penalty kicks against random, static, and adversarial goalkeepers.

\subsection{Analysis of motor dimensionality}
\label{sec:servette-dim-analysis}
The challenge of controlling over-actuated, high-dimensional musculoskeletal systems is well-documented in neuroscience~\cite{bernstein1967coordination, loeb2021learning}.
An influential hypothesis claims that brain solves this challenge through simplification by grouping functionally or anatomically related muscles together and executing control at that group level~\cite{tresch2009case}.
In animals and humans, many previous studies have used principal component analysis to show that both kinematic trajectories and electromyography signals evolve along trajectories that are low-dimensional relative to their degrees of freedom.
However, an alternative explanation for these findings is that low dimensionality is a consequence of and not a requirement for good motor control~\cite{loeb2021learning}.
To distinguish these two accounts, it is necessary to quantify control dimensionality via its effect on task performance instead of through signal reconstruction.
While such causal experiments are difficult in real animals, they are particularly well-suited for in-silico models of motor control, where the control signal can be analysed and manipulated at will in real time.

We recorded muscle activity patterns generated by the musculoskeletal policy during the penalty-kick task.
We then extracted all principal components from the 292-dimensional signal acquired by concatenating 1000 episodes.
We found that 50 components were sufficient to explain 95.03\% of the control variance, significantly lower than the full control dimensionality.
We subsequently performed a control subspace inactivation (CSI) experiment, where, at each time step, we projected the control signal produced by the policy onto a subspace spanned by a subset of PCs prior to applying it to the muscles (see Chiappa et al.~\cite{chiappa2024acquiring} for method details).
Interestingly, we found that 95\% of relative task performance was only reached with 120 PCs or more, despite these components only providing marginal explained variance.

These findings were consistent with previous studies done on dexterous hand movement, general locomotion, as well as across motor tasks performed by a single control policy~\cite{chiappa2024acquiring, simos2025kinesis, chiappa2025arnold0}, suggesting that signal reconstruction methods may underestimate the functional dimensionality of control signals. They furthermore suggest that the low-dimensionality of the control signal is a consequence of and not a requirement for good motor control~\cite{loeb2021learning}. 

\subsection{Ablation: Pipeline Stage Contributions}

The teacher policy achieved a goal rate of 25\% when evaluated in the
MyoChallenge environment. After on-policy behavior cloning, the student policy
achieved a goal rate of 34\%, yielding a 9 percentage-point improvement over the
teacher, with a final muscle activation MSE of approximately 0.019 between
student and teacher. This improvement indicates that the student not only
successfully replicated the teacher's behavior but may have benefited from operating in
the MyoChallenge observation space and with random noise.

After performing RL fine-tuning for 100K steps, the goal conversion rate remained at approximately 34\%, but energy consumption was substantially reduced. This result indicates that fine-tuning primarily serves to improve energy efficiency rather than increase the goal conversion rate. In hindsight, the reason performance did not improve further may be attributed to the provided observations, which did not include potentially relevant information, such as the goalkeeper's state. A flexible aiming mechanism might also further improve performance. 

\subsection{Compute Resources}
\label{app:compute}

All models were trained on a single NVIDIA A100 GPU and 128 CPU threads, with 128 parallel environment instances. Stage~1 training, from imitation
learning to the ball-kicking policy, took approximately 14 days.

\subsection{Training and Test Details}
\label{app:training}

KINESIS is described in Simos et al.~\cite{simos2025kinesis}.

\subsection{Code availability}
\label{app:code}

KINESIS: (including motion imitation, target reaching, and ball kicks) code is
publicly available at \url{https://github.com/amathislab/Kinesis}.

Our submission repository is publicly available at \url{https://github.com/amathislab/mc25-soccer-servette-myoclub}.

\newpage
\section{Advocacy, Tutorials, and Baseline}\label{appendix: tutorial&bl}
To broaden the reach of the competition, we recruited a global and diverse team of volunteers working with the MyoSuite toolbox to improve accessibility and participation. The current advocacy team is listed here: \url{https://sites.google.com/view/myosuite/myochallenge/advocacy-2025/current-advocates}

Outreach was conducted through multiple channels, including direct invitations to prior MyoSuite users, targeted emails to relevant research groups, announcements on community forums, and promotion via social and professional networks. This approach helped engage contributors across different time zones, expertise levels, and application domains.

Advocates contributed both technically and pedagogically: they co-developed tutorials, tested baseline solutions, identified usability bottlenecks, and iteratively refined documentation based on user feedback. Through coordination meetings and asynchronous collaboration, the team ensured that materials remained consistent, up-to-date, and aligned with participant needs.

Together, we developed a range of support resources, including Google Colab tutorials, step-by-step video guides, baseline implementations, and interactive workshops, as summarized below.

\textbf{Colab Tutorials:}
\begin{itemize}
    \item MyoChallenge Tutorial1 - Getting Started with MyoSuite:  \url{https://colab.research.google.com/drive/1AqC1Y7NkRnb2R1MgjT3n4u02EmSPem88?usp=sharing}
    \item MyoChallenge Tutorial2 - Getting Started with Table Tennis Baselines: \url{https://colab.research.google.com/drive/15w_dzEYI9SVFLFW_OmaDOwtWJ9tu4os6?usp=sharing}
    \item MyoChallenge Tutorial3 - Submission Instructions: \url{https://colab.research.google.com/drive/11vRvWMWykNrd_5ViJVGdLXz2pnbc5QEs?usp=sharing}
\end{itemize}

\textbf{Manipulation Baselines Huggingface URL:} \url{hf://cheryl-tootty/mc25-tt-baseline}

\textbf{MyoChallenge 25' Video Tutorials and Webinars:} 
\begin{itemize}
    \item MyoChallenge 25' Kick Off Webinar: \url{https://youtu.be/5u-BEDnHa0k?si=Y6BJtLdrQZFQDe1y}
    \item Reinforcement Learning for human motion control: \url{https://youtu.be/fcYkDYVRPho?si=oB6qXLyC6z5k9Wyn}
    \item Tutorial: An introduction to using MyoChallenge Bot : \url{https://youtu.be/v8PZYj0wNGw?si=Fi4ZnRICxFZZEGYu}
    \item MyoChallenge'25 Video Walkthrough by Siyuan Liu: \url{https://www.youtube.com/watch?v=4v33Nv_kMzk}
    \item MyoChallenge'25 2nd Webinar: \url{https://youtu.be/mOrr4Sf5EGc?si=vYQgkzYVpUP4MOZa}
\end{itemize}

\textbf{Documentation: }
\begin{itemize}
    \item Official MyoChallenge '25 Documentation: \url{https://myosuite.readthedocs.io/en/latest/challenge-doc2025.html}
    \item MyoChallenge '25 GitHub Discussion Board: \url{https://github.com/MyoHub/myosuite/discussions/categories/myochallenge-2025}
\end{itemize}

\newpage
\section{MyoChallenge Bot} \label{appendix:myochallenge_bot}
\begin{figure}[htbp]
    \centering
    \begin{subfigure}[b]{0.45\textwidth}
        \centering
        \includegraphics[width=\linewidth, height=3cm]{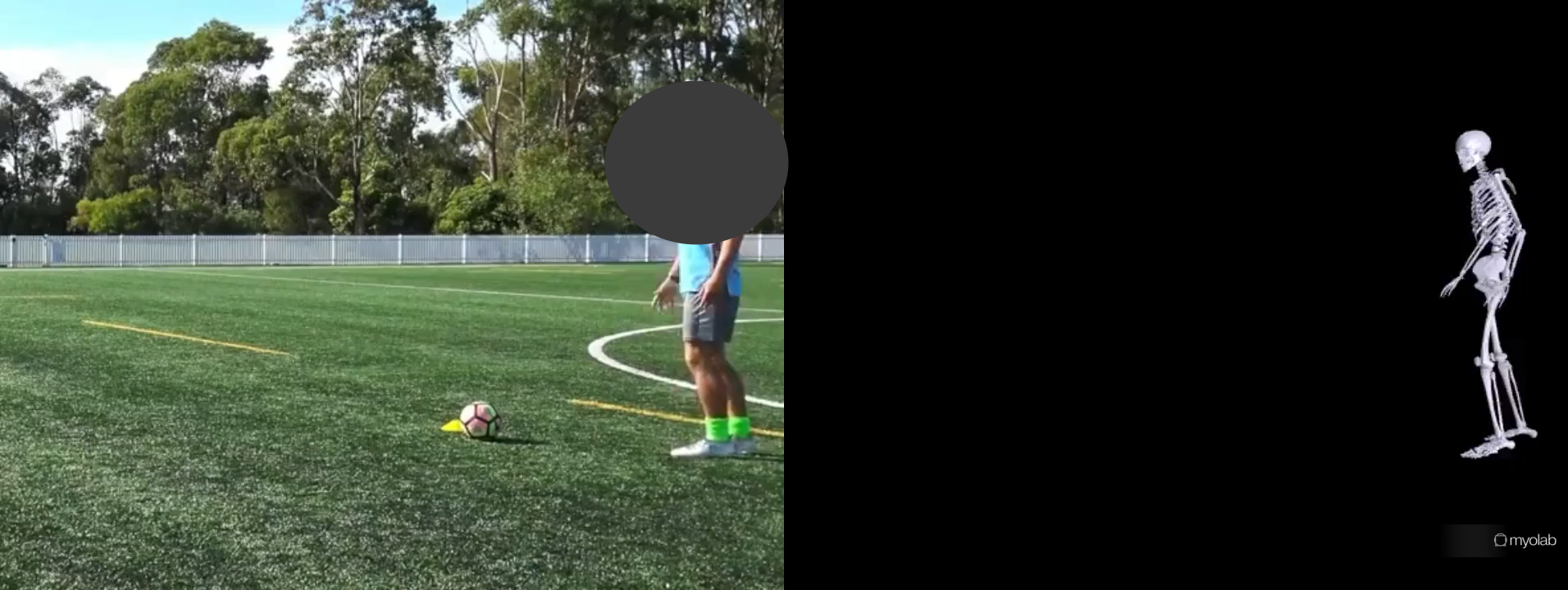}
    \end{subfigure}
    \hfill
    \begin{subfigure}[b]{0.45\textwidth}
        \centering
        \includegraphics[width=\linewidth, height=3cm]{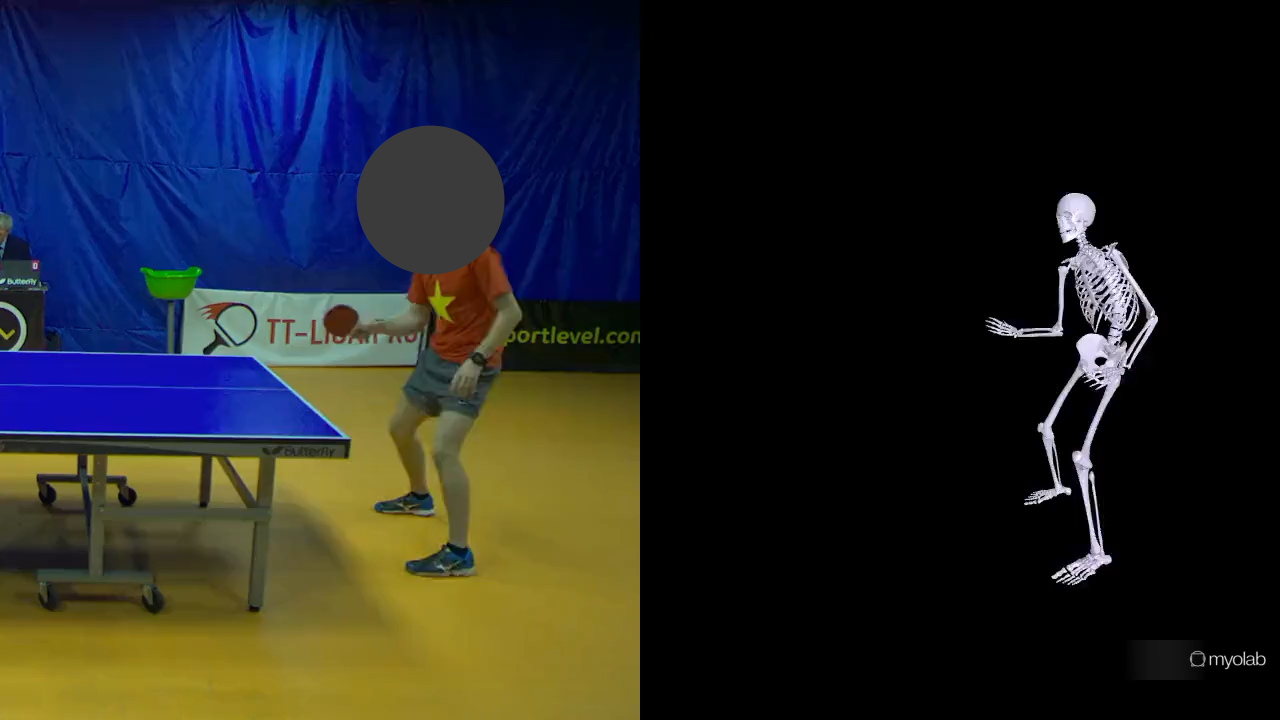}
    \end{subfigure}

    \vspace{2ex}

    \begin{subfigure}[b]{0.45\textwidth}
        \centering
        \includegraphics[width=\linewidth, height=3cm]{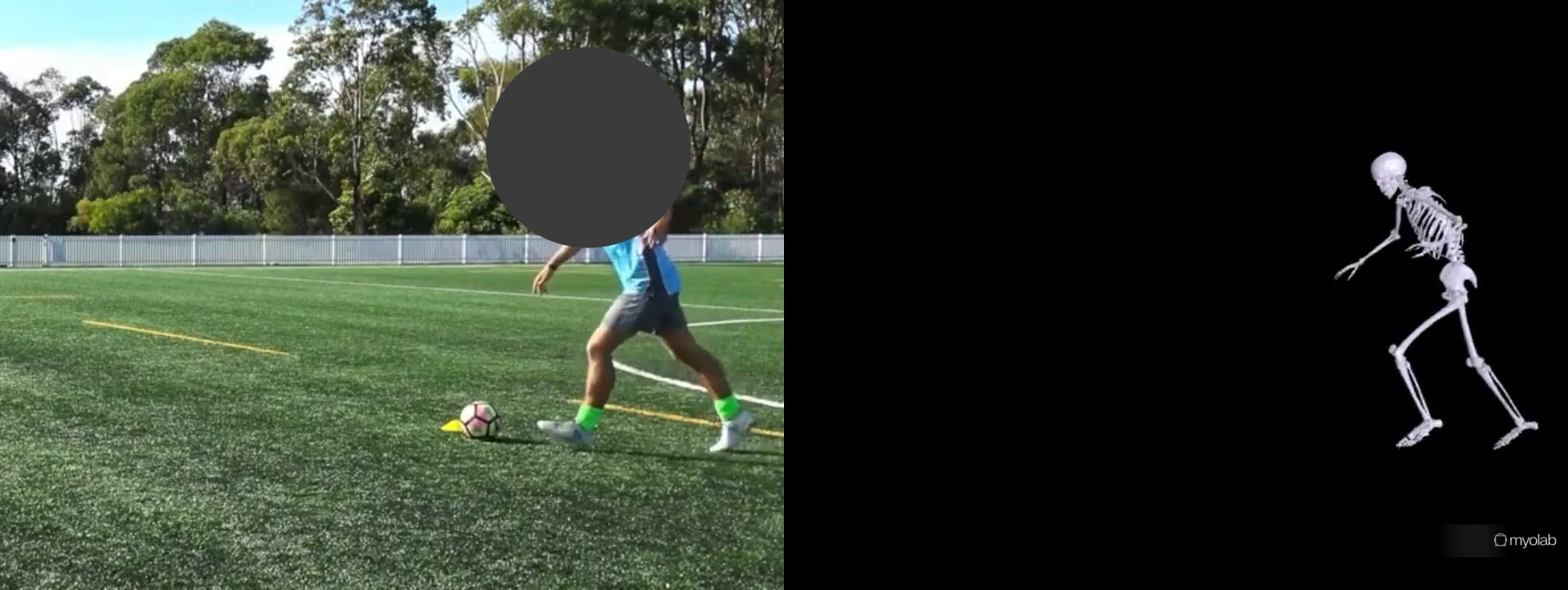}
    \end{subfigure}
    \hfill
    \begin{subfigure}[b]{0.45\textwidth}
        \centering
        \includegraphics[width=\linewidth, height=3cm]{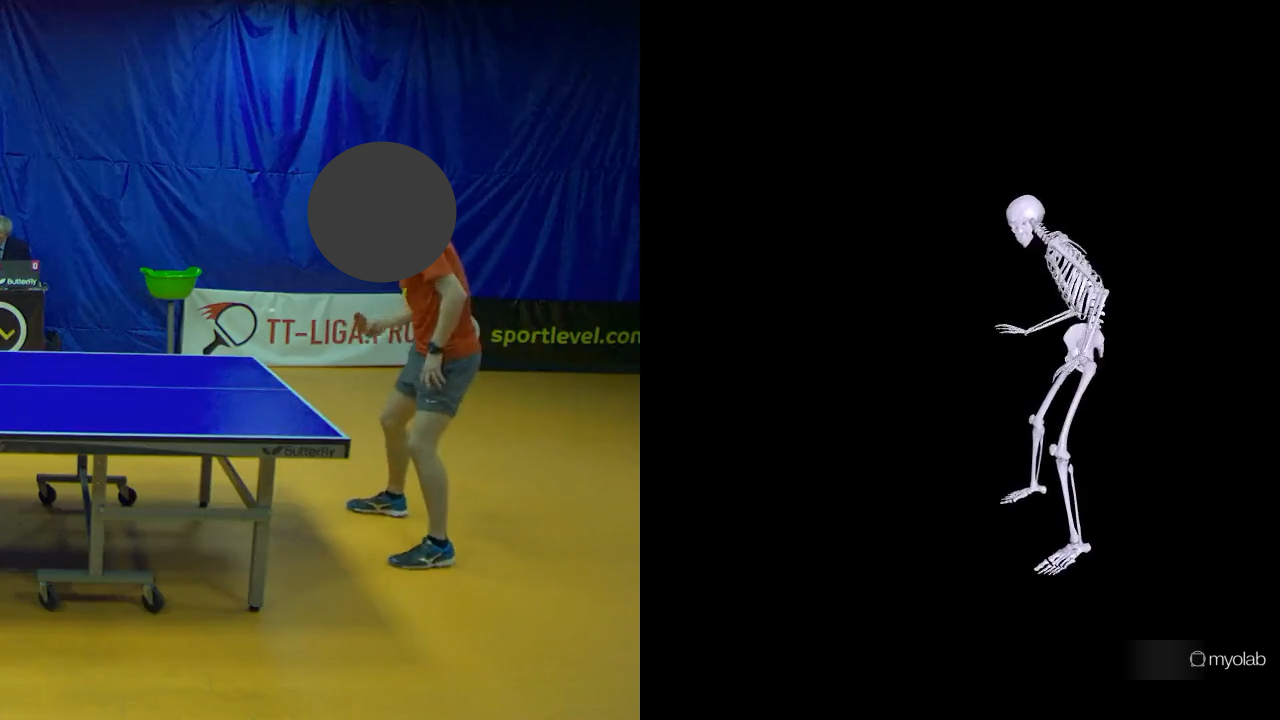}
    \end{subfigure}

    \vspace{2ex}

    \begin{subfigure}[b]{0.45\textwidth}
        \centering
        \includegraphics[width=\linewidth, height=3cm]{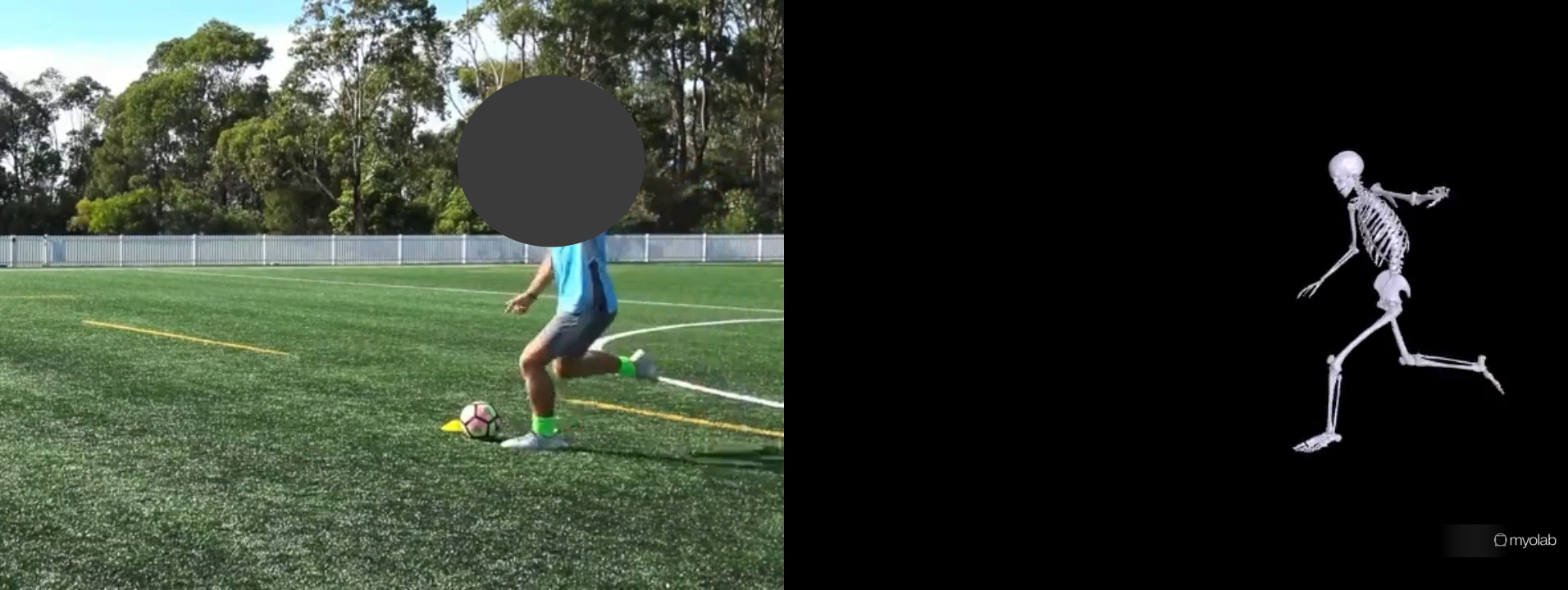}
    \end{subfigure}
    \hfill
    \begin{subfigure}[b]{0.45\textwidth}
        \centering
        \includegraphics[width=\linewidth, height=3cm]{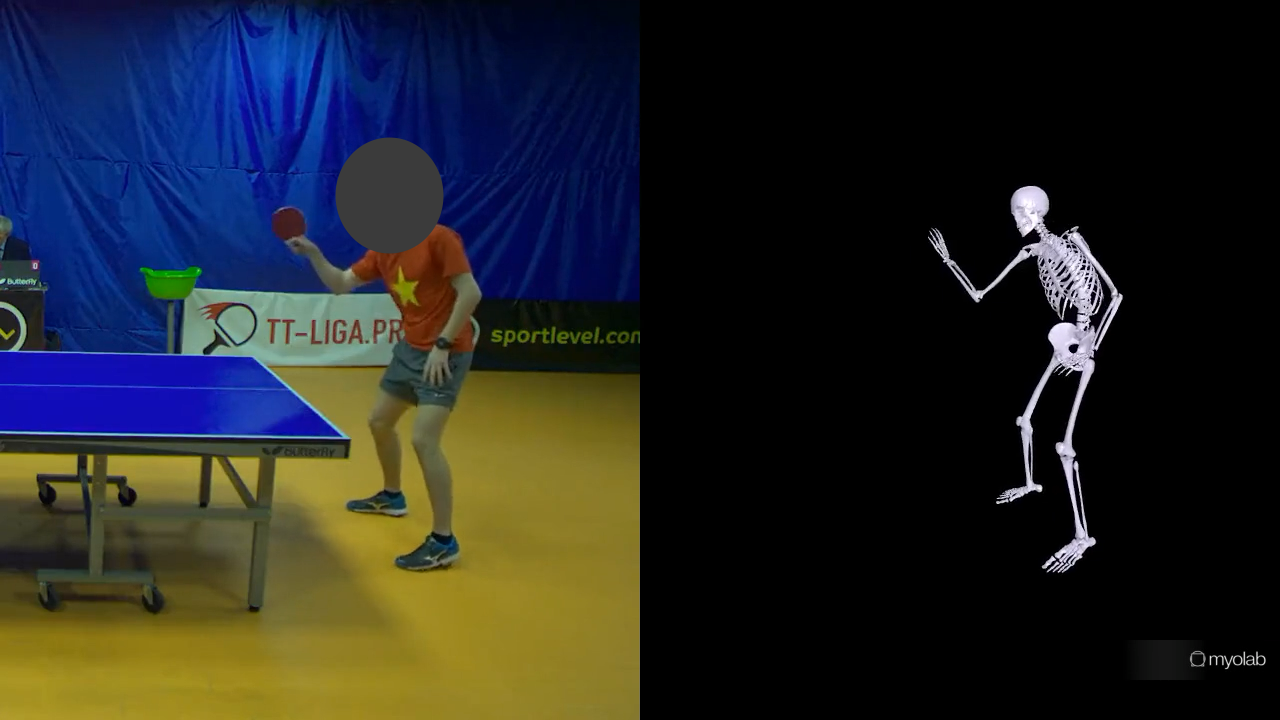}
    \end{subfigure}

    \vspace{2ex}

    \begin{subfigure}[b]{0.45\textwidth}
        \centering
        \includegraphics[width=\linewidth, height=3cm]{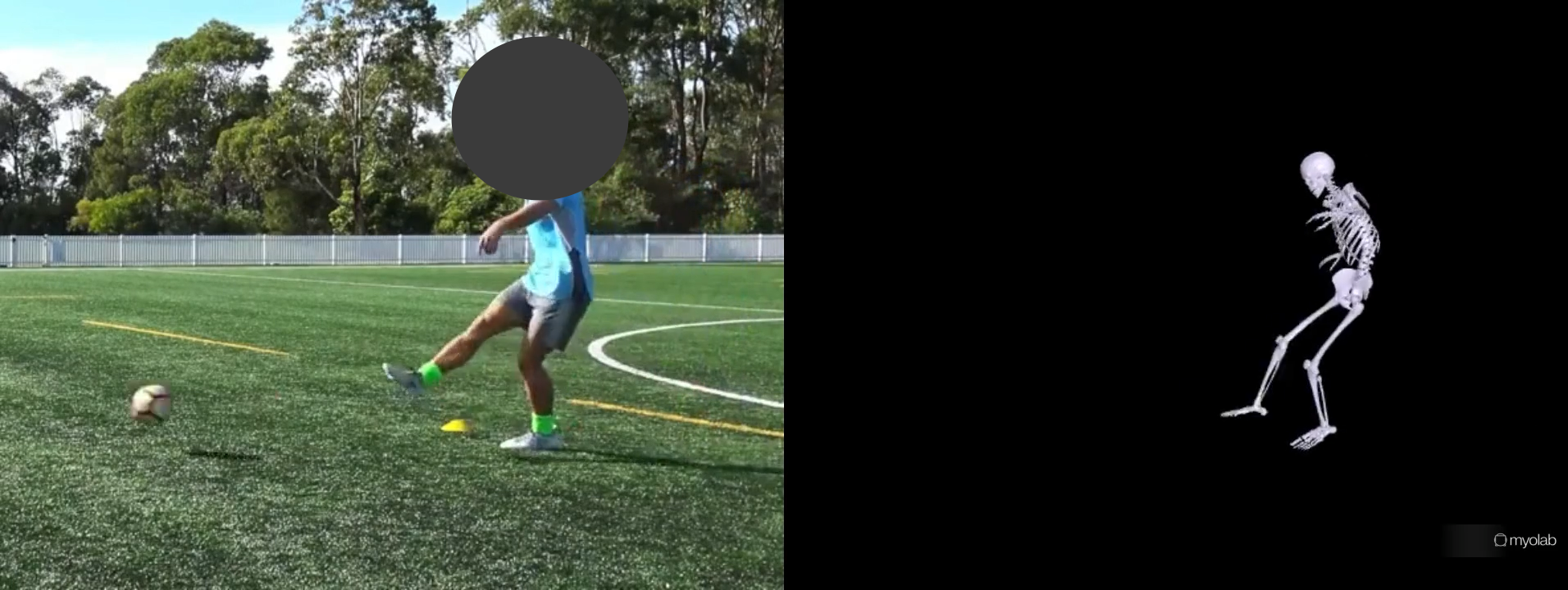}
    \end{subfigure}
    \hfill
    \begin{subfigure}[b]{0.45\textwidth}
        \centering
        \includegraphics[width=\linewidth, height=3cm]{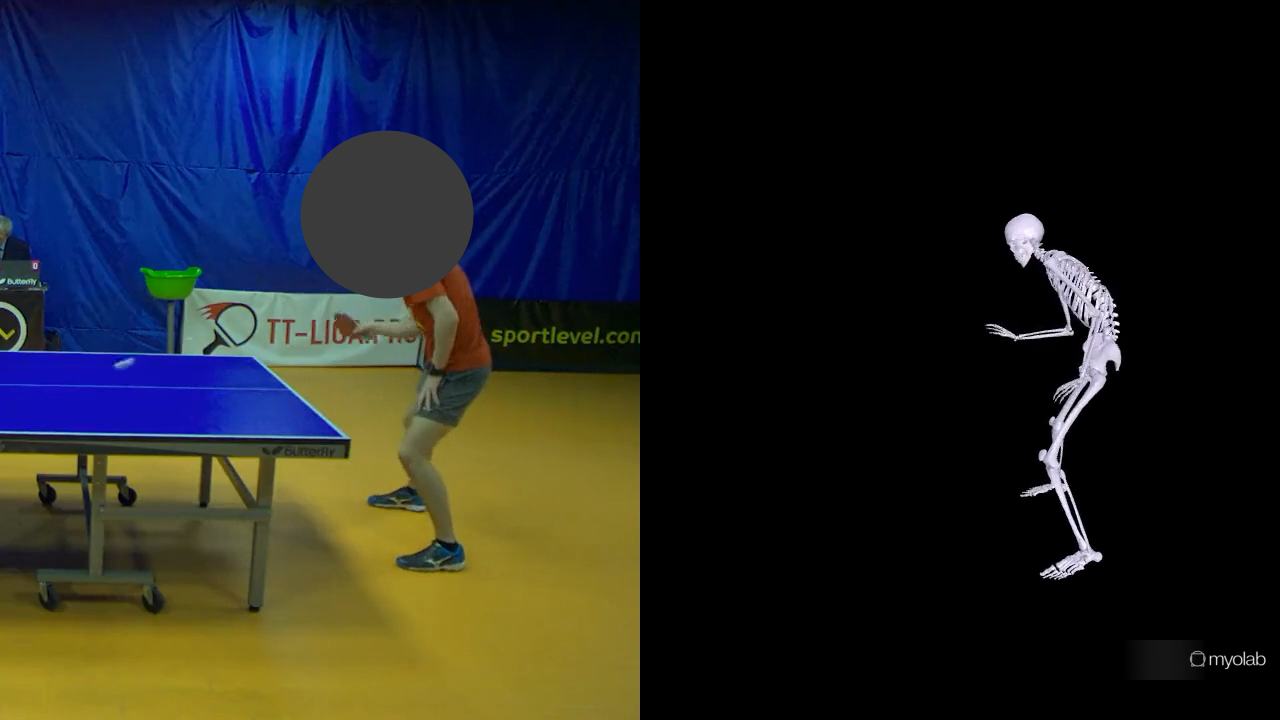}
    \end{subfigure}

    \vspace{2ex}

    \begin{subfigure}[b]{0.45\textwidth}
        \centering
        \includegraphics[width=\linewidth, height=3cm]{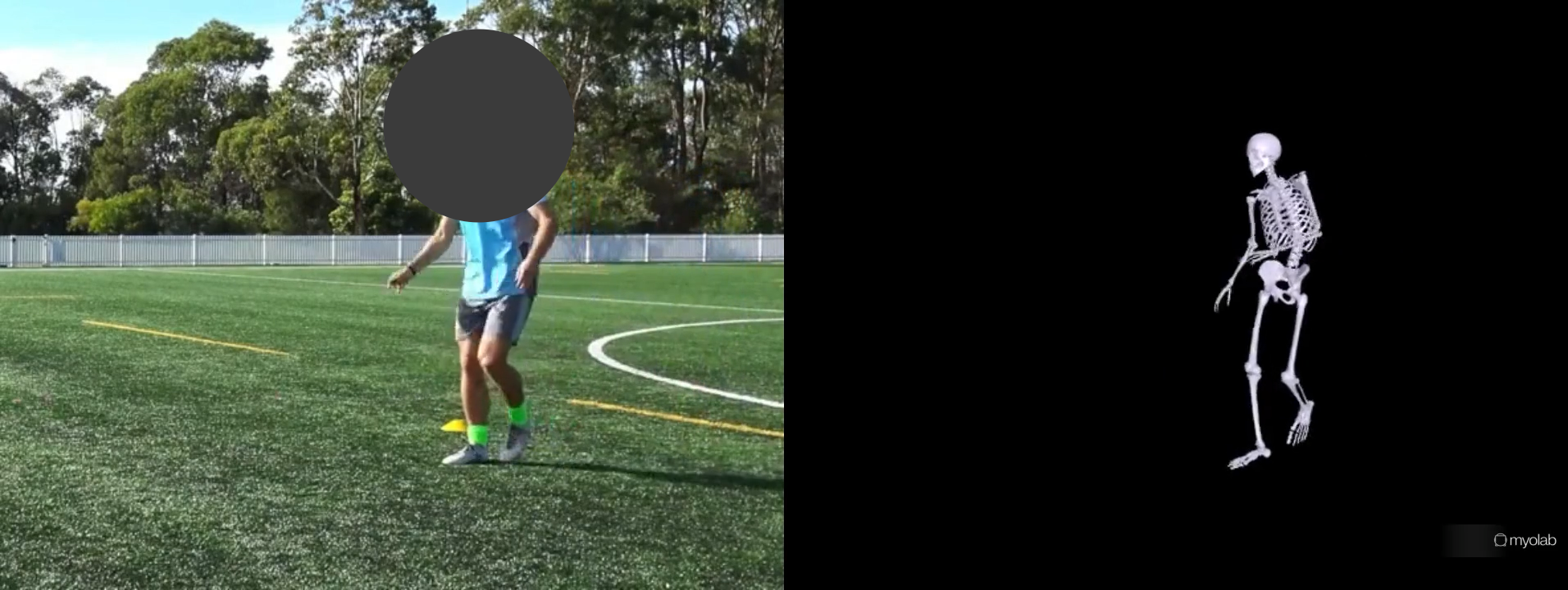}
        \caption*{A: example soccer video}
    \end{subfigure}
    \hfill
    \begin{subfigure}[b]{0.45\textwidth}
        \centering
        \includegraphics[width=\linewidth, height=3cm]{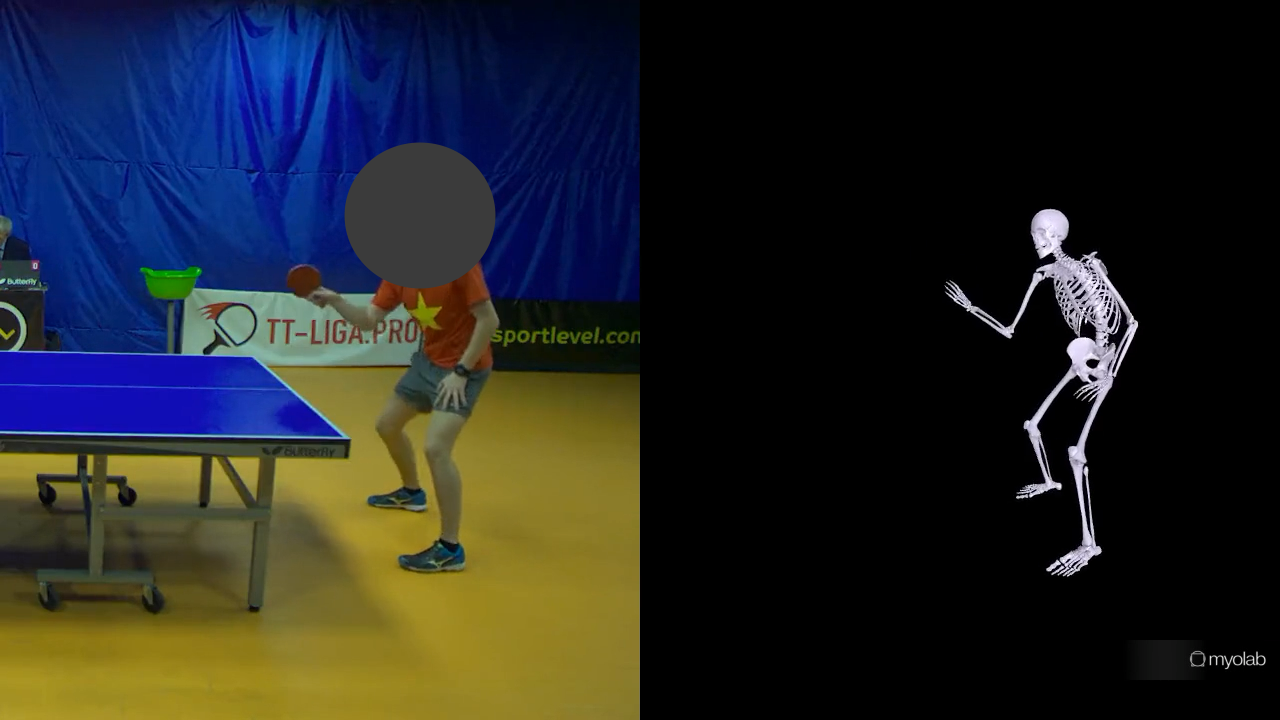}
        \caption*{B: example table tennis video}
    \end{subfigure}

    \caption{Example video conversion ability of MyoChallenge Bot.\textbf{ A:} left column as soccer penalty kick example.\textbf{ B}: right as table tennis rally example.}
    \label{fig:myochallenge_bot}
\end{figure}

In this edition of MyoChallenge, we also provide the participants with a Discord bot that allows the generation of kinematics data of the MyoSkeleton \cite{myoskeleton} from any uploaded video in the form of an h5 file and a skeleton video, as shown in Figure \ref{fig:myochallenge_bot}. Participants are able to convert the kinematics references from myoskeleton to h5 files and use imitation learning on self-recorded or internet data if necessary.

\end{document}